\documentclass[final,3p,times,authoryear]{elsarticle}

\usepackage{amssymb, amsfonts}
\usepackage{amsmath, amsthm}
\usepackage{lineno}

\usepackage{xcolor}
\usepackage{float}
\usepackage{caption}
\usepackage{listings}
\usepackage{algorithm}
\usepackage{algorithmic}
\usepackage{comment}
\usepackage{booktabs}
\usepackage{multirow}
\usepackage{subcaption}
\usepackage{xcolor}
\usepackage{proofread}
\usepackage{rotating}
\definecolor{agentagree}{RGB}{200,0,0}   
\definecolor{classagree}{RGB}{0,90,200}

\usepackage{graphicx}
\usepackage{array}

\journal{Neural Networks}

\noproofreadmark

\begin{document}

\begin{frontmatter}



\title{SimSiam Naming Game: A Unified Approach for \\ Emergent Communication and Representation Learning} 

\author[label1]{Nguyen Le Hoang}
\ead{nguyen.lehoang.4a@kyoto-u.ac.jp}
\author[label1,label2]{Tadahiro Taniguchi}
\author[label1]{Fang Tianwei}
\author[label3]{Akira Taniguchi}
\author[label1]{Masatoshi Nagano}

\affiliation[label1]{organization={Graduate School of Informatics, Kyoto University, Kyoto, Japan}}
\affiliation[label2]{organization={Research Org. of Science and Technology, Ritsumeikan University, Shiga, Japan}}
\affiliation[label3]{organization={College of Info. Science \& Engineering, Ritsumeikan University, Osaka, Japan}}

\begin{abstract}
Emergent Communication (EmCom) investigates how agents develop symbolic communication through interaction without predefined language. Recent frameworks, such as the Metropolis–Hastings Naming Game (MHNG), formulate EmCom as the learning of shared external representations negotiated through interaction under joint attention, without explicit success or reward feedback. However, MHNG relies on sampling-based updates that suffer from high rejection rates in high-dimensional perceptual spaces, making the learning process sample-inefficient for complex visual datasets. In this work, we propose the SimSiam Naming Game (SSNG), a feedback-free EmCom framework that replaces sampling-based updates with a symmetric, self-supervised representation alignment objective between autonomous agents. Building on a variational inference–based probabilistic interpretation of self-supervised learning, SSNG formulates symbol emergence as an alignment process between agents’ latent representations mediated by message exchange. To enable end-to-end gradient-based optimization, discrete symbolic messages are learned via a Gumbel–Softmax relaxation, preserving the discrete nature of communication while maintaining differentiability. Experiments on CIFAR-10 and ImageNet-100 show that the emergent messages learned by SSNG achieve substantially higher linear-probe classification accuracy than those produced by referential games, reconstruction games, and MHNG. These results indicate that self-supervised representation alignment provides an effective mechanism for feedback-free EmCom in multi-agent systems.
\end{abstract}




\end{frontmatter}

%
%


\begin{figure}[!]
\centering
\includegraphics[width=.7\textwidth]{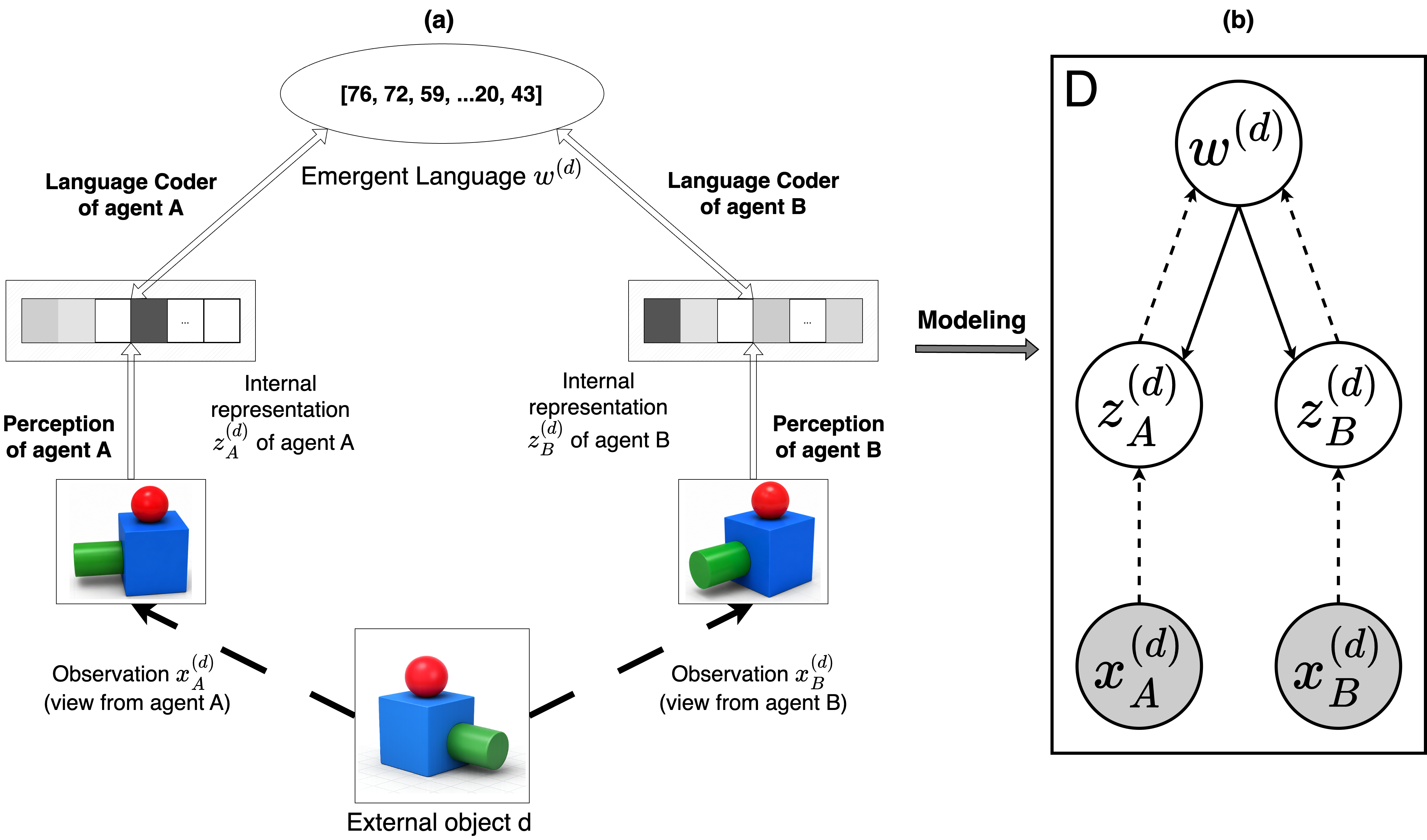}
\caption{
Two-agent interaction under the SimSiam Naming Game (SSNG).
(a) Agents A and B observe different views of the same object, encode them into latent representations, and communicate using discrete messages generated by their language coders. 
(b) Probabilistic graphical model (PGM) of SSNG. Each agent encodes its observation ($x_A^{(d)}$, $x_B^{(d)}$) into latent representations ($z_{A(d)}$, $z_{B(d)}$). Communication is mediated through a message $w_{(d)}$, which is generated from one agent’s latent state and interpreted by the other, enabling alignment of their latent representations across different views of the same underlying object. The subscript $(d)$ denotes the data index; for simplicity, this index is omitted in the equations throughout the text.
}
\label{fig: EmCom}
\end{figure}


\section{Introduction}\label{sec_introduction}
Emergent Communication (EmCom) studies how autonomous agents develop communication systems through interaction rather than relying on predefined linguistic structures \citep{1995_steels_emcom, 1998_cangelosi_emlang, 2002_kirby_emcom_ILM}. Situated at the intersection of artificial intelligence, cognitive science, and linguistics, EmCom is closely related to the framework of symbol emergence systems (SES), which investigates how symbols and emergent language (EmLang) arise as external representations through multi-agent interaction \citep{2016_taniguchi_symbol_robotics}. For effective communication, these symbols should capture the semantic structure of the objects or concepts they represent \citep{2019_taniguchi_symbol_cognitive, 2025_peters_emlang_survey}. However, developing such communication systems is challenging because agents must simultaneously learn perceptual representations, infer internal meanings, and establish shared conventions to express them \citep{2020_lazaridou_emcom_survey}. Consequently, designing mechanisms that enable agents to construct structured and informative symbolic representations remains an important challenge in EmCom research \citep{2024_boldt_emcom_application_survey}.

Language games provide a principled framework for studying how pairs of agents develop communication protocols from perceptual observations \citep{1969_lewis_signaling_game, 2012_steels_naming_game}. Prior work has shown that symbolic communication can emerge in artificial agents through settings such as referential games \citep{2017_lazaridou_referential} and reconstruction games \citep{2021_mu_emcom_reconstruction}. 
Although these settings have been widely used in EmCom research \citep{2023_brandizzi_emcom_survey}, they typically frame communication as a signal-passing process in which learning is guided by explicit task-based feedback \citep{2018_mordatch_gridworld}, such as environmental rewards \citep{2016_foerster_emcom_marl}, listener accuracy \citep{2017_havrylov_referential}, or pixel-level reconstruction loss \citep{2020_chaabouni_recon_compositionality_generalization}. In contrast, studies in developmental psychology suggest that infants acquire their earliest words through interactional cues such as joint attention, whereby learners coordinate attention with others toward shared referents \citep{1986_tomasello_joint_attention}. Motivated by this insight, the Metropolis–Hastings Naming Game (MHNG) has been proposed as a feedback-free framework in which agents negotiate symbolic categories through interaction under shared attention to their environment \citep{2019_hagiwara_inter_dm, 2023_taniguchi_inter_gmm_vae}.

While MHNG provides an elegant, interaction-driven formulation of EmCom, it updates symbolic representations through Metropolis–Hastings (MH) sampling to approximate Bayesian inference \citep{2004_robert_montecarlo}. MH algorithms rely on sampling techniques that can struggle in the high-dimensional continuous spaces typical of perceptual observations \citep{1996_mengersen_mh, 1997_gelman_mh, 2001_roberts_mh}. In EmCom settings, this often leads to excessively high rejection rates during proposal updates, making the learning process sample-inefficient and difficult to scale. This limitation motivates the development of an alternative formulation that preserves the feedback-free, interaction-driven structure of MHNG while avoiding explicit sampling, enabling agents to efficiently develop EmLang as external symbolic representations.

From a representation learning perspective, symbol emergence can be understood as a process in which agents collectively develop shared external representations through interaction while observing the same object from different viewpoints \citep{2024_taniguchi_CPC}. In this process, each agent encodes perceptual inputs into private internal representations and generates communicative symbols accordingly. This encourages the internal representations to become implicitly consistent across viewpoints, thereby enabling agents to negotiate shared symbols. This view closely parallels self-supervised learning (SSL), a widely used paradigm for representation learning that aligns representations of multiple views of the same data point in latent space \citep{2021_liu_ssl_survey, 2024_hu_contrastive_survey}. Specifically, approaches like SimSiam \citep{2021_chen_simsiam} process both network branches symmetrically via direct gradient updates. This structural symmetry makes SimSiam naturally suited for modeling two equal, autonomous agents engaged in bidirectional communication. Furthermore, a recently proposed probabilistic interpretation of SSL based on variational inference (VI) bridges self-supervised representation alignment with generative modeling \citep{2023_nakamura_visimsiam}, enabling the principled integration of SimSiam-based alignment objectives into EmCom frameworks.

Building on this connection, we propose the SimSiam Naming Game (SSNG), a feedback-free language game in which agents update their representations using a SimSiam-inspired alignment objective \citep{2021_chen_simsiam}. The two agents are implemented as independent but structurally symmetric branches of a Siamese architecture \citep{1994_bromley_siamese}, each processing its own perceptual observation without shared parameters. Each agent consists of a perception module that encodes its observation into a latent space and a language coder responsible for message generation and interpretation (Fig.~\ref{fig: EmCom}(a)). Discrete messages are produced via the Gumbel–Softmax relaxation \citep{2017_jang_categorical, 2017_maddison_categorical} with a Straight-Through estimator \citep{2013_bengio_ste}, enabling discrete communication while preserving end-to-end differentiability. Through repeated message exchanges, the two agents gradually align their latent representations and co-develop a shared communication system.

SSNG leverages self-supervised representation alignment principles \citep{2022_balestriero_ssl, 2025_uelwer_SSL_survey} to learn emergent symbolic messages that serve as external representations capturing semantic structure. To evaluate the semantic richness of the emergent language, we apply linear probing—a standard evaluation protocol in SSL—directly to the discrete messages produced by the agents. Experiments on CIFAR-10 \citep{2009_krizhevsky_cifar} and ImageNet-100 \citep{2009_deng_imagenet} show that SSNG achieves higher top-1 classification accuracy than referential games, reconstruction games, and MHNG. These results demonstrate that our self-supervised alignment objective encourages the emergence of symbols that capture highly informative, linearly separable semantic structures useful for downstream tasks.

The main contributions of this paper are as follows:
\begin{itemize}
\item We introduce the SimSiam Naming Game (SSNG), a feedback-free EmCom framework in which two autonomous agents form a shared EmLang as external representations by aligning their internal representations using a symmetric, SimSiam-inspired objective.
\item We show that merging the two independent SSNG agents into a single network recovers the performance of standard SimSiam. This demonstrates that SSNG successfully extends SimSiam’s centralized representation learning properties into a decentralized EmCom scenario without a significant loss in representation quality.
\end{itemize}

The remainder of this paper is organized as follows: Section 2 reviews related work and background; Section 3 presents the proposed framework; Section 4 reports experimental results; and Section 5 concludes with a discussion of implications and future directions.

\delspan{-We introduce the SimSiam Naming Game (SSNG), a feedback-free communication framework in which agents align their internal representations through interaction. Each agent contains a perceptual encoder and a language coder that jointly support the emergence of discrete symbolic messages.}
\delspan{-We provide a probabilistic interpretation of SSNG that views communication as an inference process over latent representations and messages. This connects SSNG to broader variational inference perspectives and clarifies its conceptual relationship to self-supervised representation alignment methods.}
\delspan{These results suggest that SSL provides a promising foundation for feedback-free EmCom and provides new insights into how semantically structured symbolic systems can develop through interaction in multi-agent settings.}
\delspan{In EmCom settings, where messages and internal representations are often high-dimensional, these sampling-based updates can lead to slow convergence, high rejection rates, and limited exploration of the communication space. As a result, MHNG faces challenges in scaling to scenarios involving longer messages, larger vocabularies, or richer perceptual representations.}
\delspan{These settings, however, leave open the question of how flexible and structured communication might emerge in the absence of such explicit feedback.}
\delspan{From a developmental perspective, early language learning is rarely driven by explicit corrective feedback. Although pointing-and-naming interactions occur, infants, around 9–15 months, acquire their earliest words primarily through joint attention, the ability to coordinate attention with another individual toward a shared referent.} 

%
%

\section{Related Work}\label{sec_preliminaries}

\subsection{Emergent Communication}\label{subsec_emcom}
EmCom studies how communication systems arise through interaction among agents and their environment rather than through predefined linguistic rules. The field has its origins in cognitive science, artificial life, and developmental robotics, where language is treated as an emergent social phenomenon. Early work demonstrated that structured communication can arise from simple interaction dynamics, including Steels’ Naming Game \citep{1995_steels_emcom, 1998_steels_mas}, evolutionary simulations of symbol emergence \citep{2003_wagner_emcom, 2002_cangelosi_emcom}, and the Iterated Learning Model (ILM) \citep{2002_kirby_emcom_ILM, 2008_kirby_human_lang}. These studies established a developmental perspective on language emergence in which shared symbolic systems arise through repeated interaction among agents. Recent surveys summarize how these ideas have been extended to modern neural architectures \citep{2020_lazaridou_emcom_survey, 2023_brandizzi_emcom_survey, 2024_suglia_VL_survey}.

\paragraph{Referential Games}
The referential game \citep{1969_lewis_signaling_game} is the most widely adopted experimental paradigm in modern neural EmCom research. In this setting, a sender observes a target and communicates a message that enables a listener to identify it among distractors. Neural implementations have shown that agents can acquire symbolic communication protocols directly from perceptual input \citep{2017_lazaridou_referential, 2018_lazaridou_referential}. Subsequent work explored recurrent messaging \citep{2017_havrylov_referential}, Transformer-based architectures \citep{2025_rita_referential}, visual perception tasks \citep{2018_bouchacourt_referential_visual}, and structured domains such as graphs \citep{2021_slowik_referential_graph}. These studies have also investigated compositionality and generalization in emergent languages \citep{2024_carmeli_referential_compositionality} and analyzed structural properties of the resulting protocols \citep{2019_chaabouni_emcom_word, 2025_kouwenhoven_referential_structure}. Referential games thus constitute a widely used baseline in EmCom research.

\paragraph{Reconstruction Games}
Reconstruction-based communication treats the emergent message as an information bottleneck similar to that of an autoencoder: the listener reconstructs the sender’s observation or latent state from the received message \citep{2020_chaabouni_recon_compositionality_generalization, 2021_mu_emcom_reconstruction}. Such settings provide a useful framework for studying how symbolic messages encode underlying structure in the data. Prior work has shown that reconstruction objectives can encourage disentangled representations and systematic symbol–meaning mappings \citep{2019_andreas_compositionality}. Reconstruction games have also been used to study compositional generalization and the trade-off between compression and expressivity in emergent languages \citep{2022_xu_compositional_generalization}. In our experiments, reconstruction games serve as a complementary baseline emphasizing information preservation.

\paragraph{Metropolis–Hastings Naming Game (MHNG)}
Recent work extends the developmental perspective of EmCom by framing communication as probabilistic inference. Collective Predictive Coding (CPC) hypothesis models interacting agents as predictive systems that minimize internal prediction error while inferring each other's latent states \citep{2024_taniguchi_CPC}, motivating the framework of \emph{generative EmCom} \citep{2025_taniguchi_generative_emcom}. Within this perspective, the Metropolis–Hastings Naming Game (MHNG) operationalizes communication as Metropolis–Hastings sampling over a shared generative model \citep{2019_hagiwara_inter_dm, 2023_taniguchi_inter_gmm_vae}. The speaker proposes a message derived from its internal state, the listener evaluates it using an acceptance ratio, and accepted proposals update both agents’ latent states. Through repeated interaction, this process converges toward a stationary distribution representing a shared lexicon. MHNG has been extended to recursive inference \citep{2023_inukai_recursive_mhng}, multimodal deep generative models \citep{2023_hoang_inter_gmm_mvae}, compositional messages \citep{2024_hoang_inter_vae_vae}, and natural-language communication between agents \citep{2025_matsui_mhcg}. Despite its flexibility, MHNG inherits the limitations of Markov Chain Monte Carlo (MCMC), including slow mixing and poor scalability in high-dimensional settings. Despite these limitations, MHNG remains an important reference point for generative and feedback-free EmCom frameworks.

\paragraph{Other Communication Frameworks}
Several alternative approaches explore communication under different assumptions. Multi-turn question–answer games \citep{2017_das_emcom_dialog, 2017_kottur_emcom_dialog, 2019_agarwal_emcom_dialog, 2023_lei_emcom_dialog} and Gridworld communication tasks \citep{2018_mordatch_gridworld} introduce iterative pragmatic reasoning but rely on explicit task feedback. In multi-agent reinforcement learning (MARL), communication channels have been used to facilitate coordination through differentiable messaging mechanisms such as DIAL \citep{2016_foerster_emcom_marl}, CommNet \citep{2016_sukhbaatar_emcom_marl}, and TarMAC \citep{2019_das_emcom_marl}. These systems are typically reward-driven and often produce continuous communication signals rather than discrete symbolic messages. Information-theoretic approaches based on the Information Bottleneck principle \citep{1999_tishby_information_bottleneck} encourage compact or compositional protocols through bottleneck constraints but rely on supervised or task-specific objectives \citep{2025_tucker_vq_vib}. While these approaches broaden the landscape of communication mechanisms, they depend on external signals such as rewards, negative samples, or reconstruction targets, components absent from our feedback-free formulation \citep{2025_galke_emlang}.

Overall, our approach is conceptually closest to generative EmCom frameworks such as MHNG, which model language formation as an inference-driven process between interacting agents. However, because these sampling-based methods suffer from sample inefficiency and high rejection rates in complex visual domains, we turn to SSL. This approach provides a robust mathematical blueprint for a more efficient communication mechanism by aligning high-dimensional continuous representations without explicit task feedback or sampling.

\subsection{Self-Supervised Representation Learning}\label{subsec_ssl}
Self-supervised learning (SSL) has become a central paradigm for learning structured representations from unlabeled data \citep{2020_jing_ssl_survey, 2025_uelwer_SSL_survey}. Many SSL methods operate by aligning representations of multiple views of the same input \citep{2020_khac_contrastive_survey, 2021_jaiswal_contrastive_survey, 2024_hu_contrastive_survey}. Contrastive frameworks such as SimCLR \citep{2020_chen_simclr}, MoCo \citep{2020_he_moco}, and CLIP \citep{2021_radford_clip, 2025_chuang_metaclip2} learn representations by maximizing agreement between positive pairs while contrasting them with negative examples.

\begin{figure}[!]
\centering
\includegraphics[width=.8\textwidth]{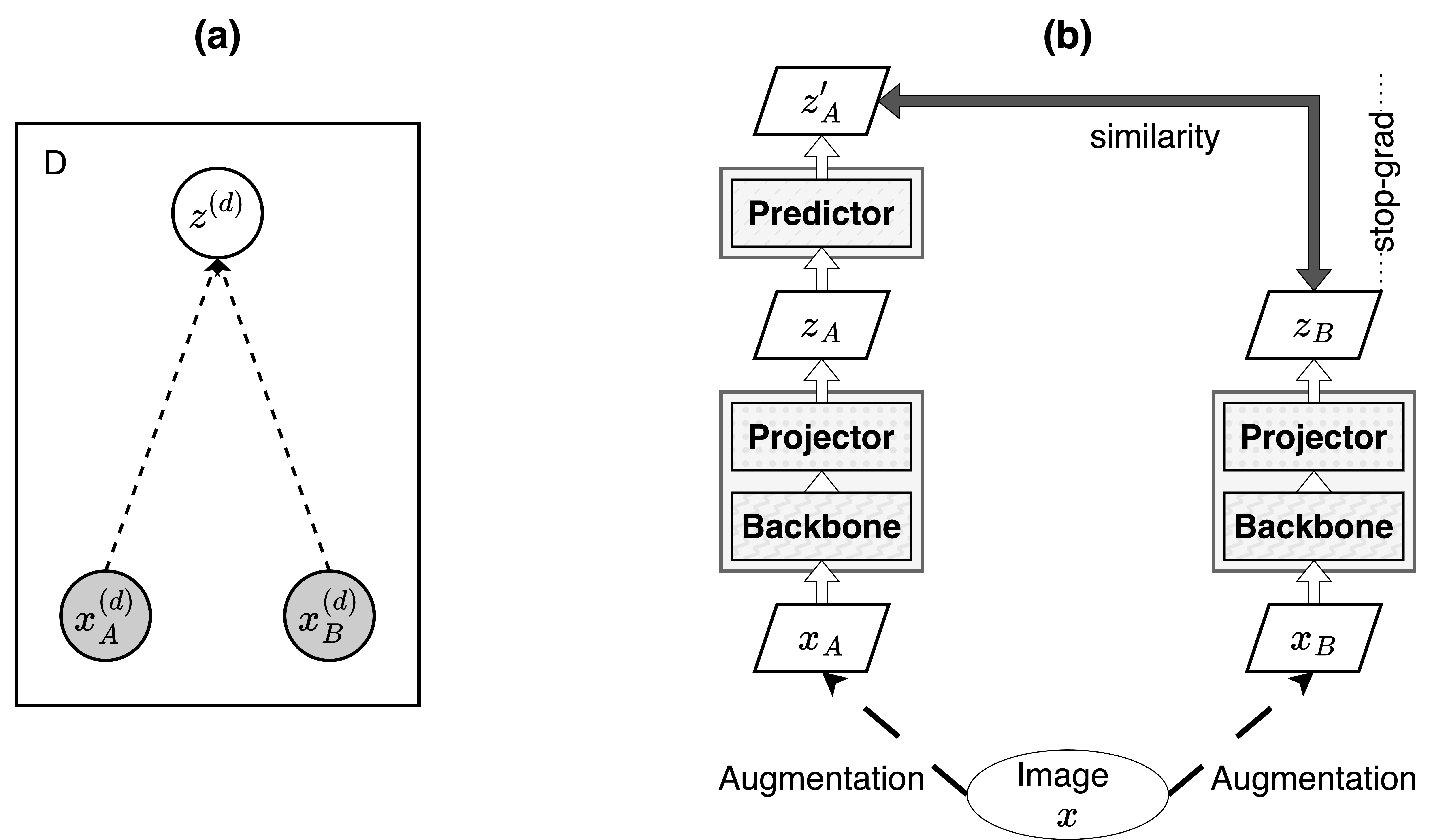}
\caption{Self-supervised representation learning interpreted from a VI perspective. 
{(a)} Probabilistic graphical model illustrating the inference process in SSL. Two augmented views $x_A$ and $x_B$ of the same data sample, drawn from dataset $D$, are treated as observations generated from a shared latent representation $z$. 
{(b)} The SimSiam framework \citep{2021_chen_simsiam}. Two augmented views are processed by a shared encoder $f$ (backbone + projector) to produce representations $z_A$ and $z_B$. A predictor $h$ transforms $z_A$ into $z'_A$, which is aligned with $z_B$ using a similarity objective. The stop-gradient operation applied to $z_B$ prevents gradient flow through the target branch, avoiding representational collapse.}
\label{fig: SSNG_SSL}
\end{figure}

A second family of SSL methods removes the need for explicit negatives. Approaches such as BYOL \citep{2020_grill_byol}, SimSiam \citep{2021_chen_simsiam} (Fig.~\ref{fig: SSNG_SSL} (b)), Barlow Twins \citep{2021_zbontar_barlowtwins}, and VICReg \citep{2022_bardes_vicreg} avoid representational collapse through architectural asymmetry or variance and covariance regularization. Teacher–student methods such as DINO and DINOv2 \citep{2021_caron_dino, 2024_oquab_dinov2} further demonstrate that predictive alignment in latent space can yield stable and expressive representations. This predictive perspective is formalized in JEPA \citep{2022_LeCun_jepa} and its extensions \citep{2024_riou_stemjepa, 2025_assran_vjepa2}, which show that latent-space prediction alone can produce scalable and robust representation learning systems. Recent theoretical work by Nakamura et al.\ \citep{2023_nakamura_visimsiam} provides a unifying probabilistic perspective by interpreting several SSL objectives as performing variational inference in latent space \citep{2017_blei_variational_inference} (Fig.~\ref{fig: SSNG_SSL} (a)).

Our work connects this line of research with EmCom. In the proposed framework, the two communicating agents are implemented as two branches of a Siamese-style architecture without shared parameters \citep{1994_bromley_siamese}. Each agent processes its own perceptual observation while exchanging discrete messages generated by its language coder. A SimSiam-inspired predictive objective aligns latent representations inferred from perceptual observations and received messages. Through repeated interaction, the agents progressively align their internal representations while simultaneously developing a shared symbolic communication system. This formulation establishes a direct connection between SSL and EmCom, enabling feedback-free language emergence.


%
%
%

\section{SimSiam Naming Game (SSNG)}\label{sec_ssng}
In this section, we formalize the SimSiam Naming Game (SSNG). We first describe the architecture of an individual agent and its core components. We then derive the SSNG learning objective for a two-agent interaction from a variational inference perspective. Next, we describe the communication process through which agents exchange messages and align their latent representations. The complete training procedure is summarized in Algorithm~\ref{alg: SSNG}. Finally, we show that when the two-agent formulation is interpreted as a single integrated system, the framework reduces to a SimSiam-like model for self-supervised representation learning, summarized in Algorithm~\ref{alg: SSNG-repLrn}. 

\subsection{Agent Structure}\label{subsec_1agent}
\begin{figure}[h]
\centering
\includegraphics[width=.7\textwidth]{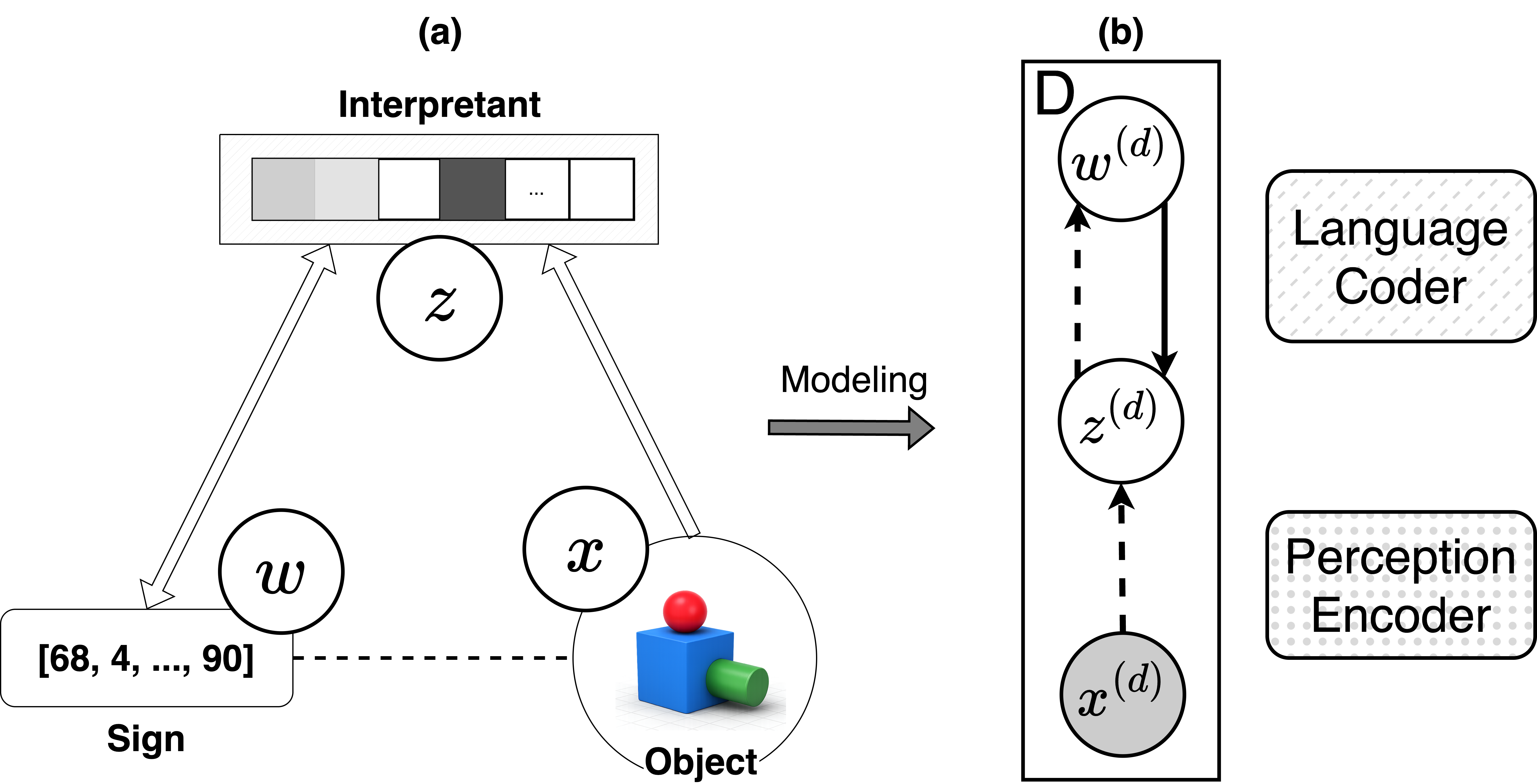}
\caption{
Agent architecture grounded in Peirce’s triadic relation. 
(a) Semiotic perspective: the {object} $x$, {sign} $w$, and {interpretant} $z$ form a triadic relation. 
Perception maps objects to the interpretant, while the language coder links the interpretant and the sign.
(b) Probabilistic model representation: the {language coder} is bidirectional, supporting both message interpretation ($w \rightarrow z$) and message generation ($z \rightarrow w$), while the {perception} operates only in the inference direction ($x \rightarrow z$). The subscript $(d)$ denotes the data index; for simplicity, this index is omitted in the equations throughout the text.
}
\label{fig:1agent}
\end{figure}

The architecture of each agent is inspired by Peirce’s triadic view of meaning: object, sign, and interpretant \citep{2002_chandler_semiotic}. In our setting (Fig.~\ref{fig:1agent}), the object corresponds to an underlying entity $x$, while each agent $i$ observes its own perceptual view $x_i$. The interpretant is the latent representation $z_i$ derived from this observation, and the sign is the message $w_i$ generated from that latent state. Each agent $i$ contains two complementary modules that operate in a common latent space: a perception encoder that produces a latent representation of the observed input,  and a language coder that maps latent representations to discrete messages and vice versa.

\paragraph{Perception Encoder}
This component maps an observed input $x_i$ into a latent representation,
\begin{equation}
z_i^{(x_i)} = f_i(x_i)
\label{eq: agent_perc_enc}
\end{equation}
where $f_i$ denotes the perceptual module of agent $i$. This produces the agent’s perceptual latent state $z_i^{(x_i)}$, which reflects the agent’s internal understanding of its observation.

\paragraph{Language Coder}
The language coder provides a bidirectional interface between latent representations and discrete messages. It consists of a message generator, which produces a sign from the agent’s latent state, and a message interpreter, which infers a latent representation from any incoming message:
\begin{align}
w_i &= h_i(z_i^{(x_i)})
\label{eq: agent_lang_enc} \\
z_i^{(w_j)} &= l_i(w_j)
\label{eq: agent_lang_dec}
\end{align}
where $h_i$ and $l_i$ denote the generator and interpreter of agent $i$, respectively. Here, $z_i^{(w_j)}$ denotes the latent representation inferred by agent $i$ from the message $w_j$, which may originate from itself $(j=i)$ or from a partner agent $(j\neq i)$.

All mappings $f_i$, $h_i$, and $l_i$ produce vectors in a shared latent space, ensuring that representations derived from perception and those inferred from messages are directly comparable. ~Although the implementation contains multiple inference pathways, each ultimately estimates the same underlying latent state $z_i$, representing the agent’s internal understanding of the observed object. This shared embedding structure is crucial, as it enables the self-supervised alignment mechanism that supports EmCom in SSNG.


\subsection{Two-Agent Interaction and Learning Objective}\label{subsec_2agent}
In SSNG, two agents interact by exchanging messages that allow them to gradually align their internal latent representations. Figure~\ref{fig: EmCom}(a) illustrates the interaction process, while Fig.~\ref{fig: EmCom}(b) presents the corresponding probabilistic graphical model that summarizes the relationships between observations, latent representations, and the communicated message. Since an agent’s internal representation cannot be directly observed by others, communication must occur through external messages that are interpreted within each agent’s latent space. We consider two agents indexed by $i \in \{A, B\}$, and denote the partner agent by $j \neq i$. Given a shared object $x$, agent $i$ observes its own perceptual view $x_i$ and encodes it into a latent representation $z_i^{(x_i)}$ using the perception encoder defined in Section~\ref{subsec_1agent}. The agent then generates a message $w_i = h_i(z_i^{(x_i)})$, which is transmitted to the partner agent.

Recent theoretical work by Nakamura et al.\ has shown that a broad class of SSL methods, including SimCLR, SimSiam, BYOL, and DINO, can be interpreted as estimating a unified latent posterior from multiple observed views of the same underlying entity \citep{2023_nakamura_visimsiam}. Under this interpretation, each network branch produces a view-specific latent that is aligned toward a shared internal representation. Motivated by this perspective, we adopt the modeling assumption that the linguistic variable $w$ represents an underlying symbolic structure shared across agents. The agent-specific messages $w_i$ and $w_j$ are treated as view-dependent estimates of this latent symbolic variable, derived from their respective perceptual representations. The SSNG alignment objective encourages these estimates to converge, in a manner analogous to how SSL methods align representations across augmented views. This provides a principled foundation for interpreting SSNG as converging toward a single emergent symbolic structure from multi-agent observations.

Building on this interpretation, the interaction mechanism of SSNG closely parallels SimSiam \citep{2021_chen_simsiam}: two views (here, the agents' perceptual inputs) are processed by separate branches whose outputs are aligned via a stop-gradient objective. In SSNG, the exchanged messages serve as a learnable communication bottleneck that mediates this alignment, extending SimSiam's two-view SSL formulation from single-image augmentations to multi-agent communication.

This interaction mechanism can also be interpreted from a probabilistic perspective. In particular, SSNG can be formulated as maximizing an evidence lower bound (ELBO) on the joint distribution $p_{\theta}(w, \mathbb{Z} \mid \mathbb{X})$, where $\mathbb{X} = \{x_A, x_B\}$ denotes the agents’ perceptual inputs and $\mathbb{Z} = \{z_A, z_B\}$ denotes their latent representations. The latent variable $w$ represents the shared symbolic structure toward which the agent-specific messages converge. To approximate the intractable posterior $p_{\theta}(w, \mathbb{Z} \mid \mathbb{X})$, we introduce a variational distribution $q_{\phi}(w, \mathbb{Z} \mid \mathbb{X})$. The parameters $\theta = \{\theta_A, \theta_B\}$ and $\phi = \{\phi_A, \phi_B\}$, corresponding to generative and inference components respectively, are optimized by maximizing the ELBO:
\begin{equation}
\theta^*, \phi^* = \arg \max_{\theta, \phi} \; \mathbb{E}_{q_{\phi}(w, \mathbb{Z} \mid \mathbb{X})} \left[ 
\log \frac{p_{\theta}(w, \mathbb{Z}, \mathbb{X})}{q_{\phi}(w, \mathbb{Z} \mid \mathbb{X})} 
\right]
\end{equation}

\noindent This formulation leads to the following objective function:
\begin{align}
\mathcal{J}_{\text{SSNG}} \simeq \mathcal{J}_{\text{align}} + \mathcal{J}_{\text{self}} + \mathcal{J}_{\text{sign}} + \mathcal{J}_{\text{uniform}}
\label{eq: loss_ssng_general}
\end{align}
where:
\begin{align}
\mathcal{J}_{\text{align}} &:= \mathbb{E}_{q_{\phi}(\mathbb{Z} \mid \mathbb{X})} \left[ \log p_{\theta}(\mathbb{Z} \mid \mathbb{X}) - \log q_{\phi}(\mathbb{Z} \mid \mathbb{X}) \right]
\label{eq: loss_ssng_align} \\
\mathcal{J}_{\text{self}} &:= \mathbb{E}_{q_{\phi}(\mathbb{Z} \mid \mathbb{X})} \left[ \mathbb{E}_{q_{\phi}(w \mid \mathbb{Z})} \left[ \log p_{\theta}(\mathbb{Z} \mid w) \right] \right]
\label{eq: loss_ssng_self} \\
\mathcal{J}_{\text{sign}} &:= - \mathbb{E}_{q_{\phi}(\mathbb{Z} \mid \mathbb{X})} \left[ D_{\text{KL}} \left( q_{\phi}(w \mid \mathbb{Z}, \mathbb{X}) \parallel p(w) \right) \right] 
\label{eq: loss_ssng_sign} \\
\mathcal{J}_{\text{uniform}} &:= \mathbb{E}_{q_{\phi}(\mathbb{Z} \mid \mathbb{X})} \left[ - \log p_D(\mathbb{Z}) \right]
\label{eq: loss_ssng_uniform}
\end{align}
The derivation is provided in~\ref{appendix_ssng_objective_function}. Each term plays a distinct role in optimizing the SSNG model:
\begin{itemize} 
    \item Alignment Objective (\(\mathcal{J}_{\text{align}}\)): Encourages alignment between an agent’s latent representation and the latent inferred from its partner’s message. 
    \item Self-Consistency Objective (\(\mathcal{J}_{\text{self}}\)): Enforces consistency between an agent’s latent representation and the latent inferred from its own message. 
    \item Sign Objective (\(\mathcal{J}_{\text{sign}}\)): Regularizes the predicted message distribution toward the prior \(p(w)\). \addspan{Intuitively, this term pulls the signs inferred by different agents closer together during interaction, promoting alignment and the emergence of a shared symbolic space.}
     \item Uniformity Objective (\(\mathcal{J}_{\text{uniform}}\)): Promotes a uniform distribution of latent states $\mathbb{Z}$ to prevent representational collapse, $p_D(\mathbb{Z})$ denotes the empirical distribution from the data. 
\end{itemize}

Following recent variational interpretations of SSL \citep{2023_nakamura_visimsiam}, we model both the sign variable $w$ and the latent variables $z_i$ as directional variables residing on hyperspheres, with uniform priors $p(w)$ and $p(z_i)$, respectively. The variational distribution over latent variables $\mathbb{Z}$ factorizes across two agents (i.e., $M=2$) and is modeled using von~Mises--Fisher (vMF) distributions. The variational distribution over the emergent message, $q_{\phi}(w \mid \mathbb{Z})$, is defined as a mixture-of-experts (MoE) in which each agent proposes a candidate message $h_i(z_i)$. This mixture form captures the idea that each agent provides a view-dependent estimate of the underlying symbolic variable, and the mixture aggregates these estimates into a unified posterior over the emergent message $w$.
\begin{align}
p(w) &:= \mathcal{U}(S^{d_w-1}) 
\label{eq: ssng_w_prior} \\
p(z_i) &:= \mathcal{U}(S^{d_z-1}) 
\label{eq: ssng_z_prior} \\
q_{\phi}(w \mid \mathbb{Z}) &:= \frac{1}{M} \sum_{i=1}^{M} \mathrm{vMF}(w ; \mu_i^{(w)} = h_i(z_i), \kappa^{(w)})  
\label{eq: ssng_w_variational} \\
q_{\phi}(\mathbb{Z} \mid \mathbb{X}) &:= \prod_{i=1}^{M} \mathrm{vMF}(z_i ; \mu_i^{(q_z)} = f_i(x_i), \kappa^{(q_z)}) 
\label{eq: ssng_z_variational} \\
p_\theta(\mathbb{Z} \mid \mathbb{X}) &:= \prod_{i=1}^{M} \left[ \mathrm{vMF}(z_i ; \mu_i^{(p_z)} = g_i(x_j), \kappa^{(p_z)}) \right]
\label{eq: ssng_z_posterior}
\end{align}
where
\begin{itemize}
\item $\mathcal{U}(S^{d_w-1})$ and $\mathcal{U}(S^{d_z-1})$ denote the uniform distributions over the unit hyperspheres $S^{d_w-1}$ and $S^{d_z-1}$, where $d_w$ and $d_z$ denote the dimensionalities of the message and latent spaces, respectively.
\item $g_i(x_j) = l_i \big( h_j( f_j(x_j) ) \big)$ denotes the compositional mapping in which the perceptual input $x_j$ from agent $j \neq i$ is encoded by $f_j$, transformed into a message by $h_j$, and subsequently interpreted in the latent space of agent $i$ by $l_i$.
\item $\mathrm{vMF}(z; \mu, \kappa) := C_{\mathrm{vMF}}(\kappa) \exp(\kappa \mu^\top z)$ is the von Mises--Fisher distribution over unit vectors $z$, with mean direction $\mu$, concentration parameter $\kappa \in \mathbb{R}^+$, and normalization constant $C_{\mathrm{vMF}}(\kappa)$  \citep{2000_mardia_statistics, 2018_davidson_svae}. All mean directions satisfy $\|\mu_i^{(w)}\|=\|\mu_i^{(q_z)}\|=\|\mu_i^{(p_z)}\|=1$. The concentration parameter $\kappa$ controls the spread: as $\kappa \to 0$, $\mathrm{vMF}$ approaches the uniform distribution on the unit hypersphere; as $\kappa \to \infty$, it collapses around $\mu$.
\end{itemize}

\paragraph{Learning Objective}
We consider a two-agent system consisting of agents $A$ and $B$. At each interaction step, both agents simultaneously process their observations and exchange messages. The learning objective is derived from the perspective of agent $i \in \{ A, B \}$, which receives message $w_j$ from its partner agent $j \neq i$. By symmetry, the same objective applies to both agents.

When processing its observation, agent $i$ computes two latent estimates from its own generated message $w_i$ and the received message $w_j$:
\[
z_i^{(w_i)} = l_i(w_i), \qquad
z_i^{(w_j)} = l_i(w_j).
\]
These reconstructions are compared against agent $i$'s perceptual latent
\[
z_i^{(x_i)} = f_i(x_i).
\]

During communication, agent $i$ adopts a message-induced local prior over $w$, centered at the received symbol from agent $j$. This encodes the assumption that the partner's message provides a prior estimate of the shared symbolic variable, and is defined as a von Mises–Fisher (vMF) distribution:
\begin{equation}
p_i(w) := \mathrm{vMF}\bigl(w;\,\mu = w_j,\,\kappa_{\text{prior}}\bigr)
\label{eq:ssng_w_prior}
\end{equation}

The ELBO formulation in Eq.~\eqref{eq: loss_ssng_general} defines the global training objective for the two-agent system. In practice, this objective is optimized from the perspective of each agent, with parameter updates derived from a per-agent contribution conditioned on the received message and its own observation. Since both message variables $w$ and latent variables $z$ lie on unit hyperspheres and are modeled using vMF distributions, the corresponding log-likelihood and KL terms reduce to scaled cosine similarity terms. Using single-sample Monte Carlo estimates and dropping additive constants yields the following per-agent objective:
\begin{align}
\mathcal{J}_i
&:=
\underbrace{
\alpha \, (z_i^{(w_j)})^\top (z_i^{(x_i)})
}_{\text{Individual Regularization}}
+
\underbrace{
\beta \, (z_i^{(w_i)})^\top (z_i^{(x_i)})
}_{\text{Individual Prediction Error}}
+
\underbrace{
\gamma \, (w_j)^\top (w_i)
}_{\text{Collective Regularization}}
\label{eq: loss_ssng_agent}
\end{align}
where $\alpha$, $\beta$, and $\gamma$ are scalar hyperparameters that control the relative importance of the three components. The derivation is provided in~\ref{appendix_ssng_elbo_listener}, and the full training objective is obtained by summing over both agents:
\begin{equation}
\mathcal{J}_{\text{SSNG}} = \mathcal{J}_A + \mathcal{J}_B
\end{equation}

Each term in Eq.~\eqref{eq: loss_ssng_agent} corresponds directly to a core component of the CPC hypothesis \citep{2025_taniguchi_cpc_science}:
\begin{itemize}
\item \textbf{Collective Regularization:} Encourages consistency between agent $i$'s emitted message $w_i$ and the message $w_j$ received from its partner, promoting the emergence of a shared communication system.
\item \textbf{Individual Prediction Error:} Measures the agreement between agent $i$'s interpretant $z_i^{(x_i)}$ and the latent reconstruction $z_i^{(w_i)}$ obtained by decoding its \emph{own} message, enforcing consistency between internal representations and emitted signs.
\item \textbf{Individual Regularization:} Measures the agreement between agent $i$'s interpretant $z_i^{(x_i)}$ and the latent reconstruction $z_i^{(w_j)}$ inferred from the \emph{partner's} message, enforcing accurate interpretation of received messages.
\end{itemize}

SSNG naturally implements SimSiam’s stop-gradient operation through the multi-agent communication channel. When Agent $j$ generates a message and sends it to Agent $i$, the message tensor is detached from Agent $j$’s computational graph. Consequently, Agent $i$ treats the incoming message purely as a fixed target for its asymmetric predictor. No gradients flow backward from Agent $i$’s loss into Agent $j$’s parameters. This detachment ensures that each agent optimizes its representations independently based on the received message, preserving the asymmetric gradient flow required for stable SimSiam optimization while maintaining the autonomy of the agents. This also ensures that Agent $i$ cannot observe or directly impact its partner's internal state.


\begin{figure}[!]
\centering
\includegraphics[width=\textwidth]{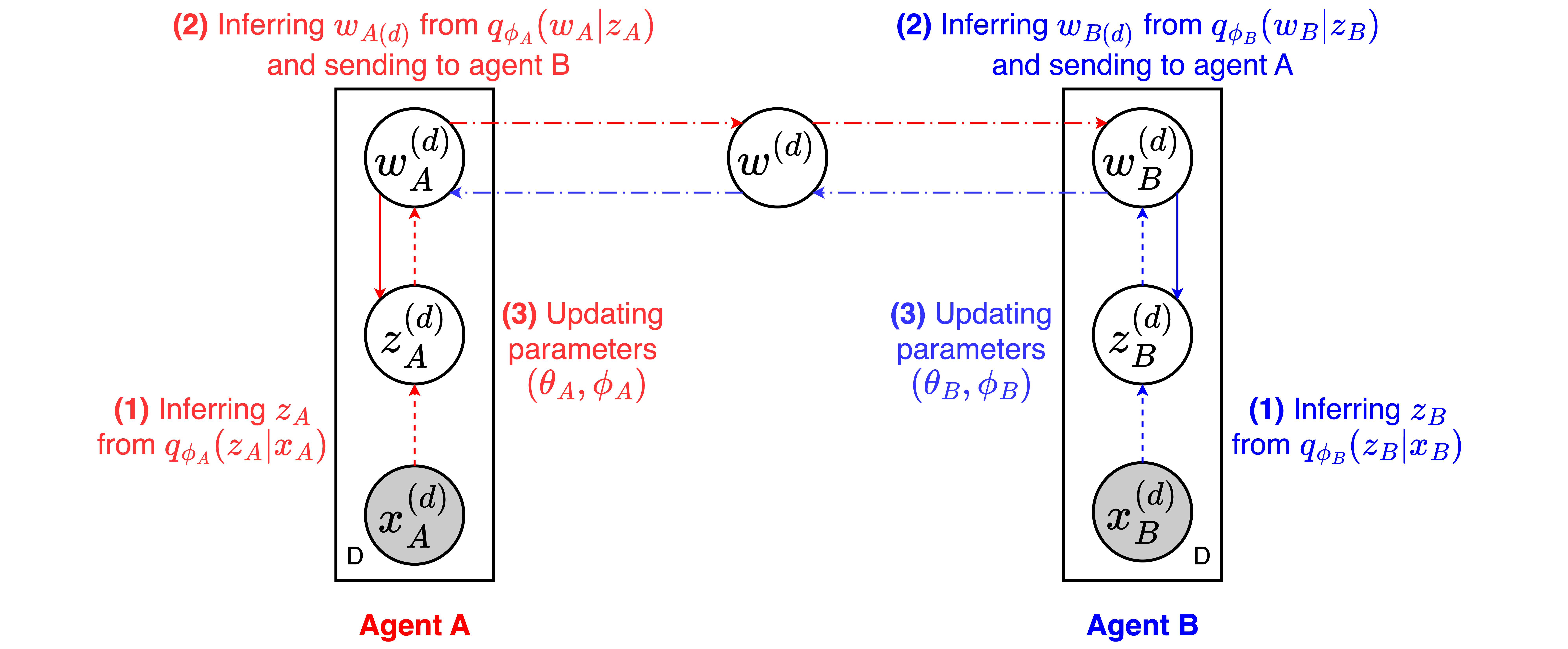}
\caption{SSNG between two agents A and B.
Agent A (red) and B (blue) perceive different views of the same object ($x_A$, $x_B$), encode them into internal latent representations ($z_A$, $z_B$), and generate messages ($w_A$, $w_B$) via their language coders. Each agent interprets the received message into a latent representation and updates its own parameters by minimizing the listener loss (Eq.~\ref{eq: loss_ssng_agent}). 
Steps (1)--(3) illustrate the symmetric inference and communication processes performed by both agents. Through repeated interaction, the agents gradually align their latent representations and co-develop a shared EmLang. The subscript $(d)$ denotes the data index corresponding to the observed object.
}
\label{fig: SSNG_proc}
\end{figure}


\begin{figure}[!]
\centering
\includegraphics[width=.6\textwidth]{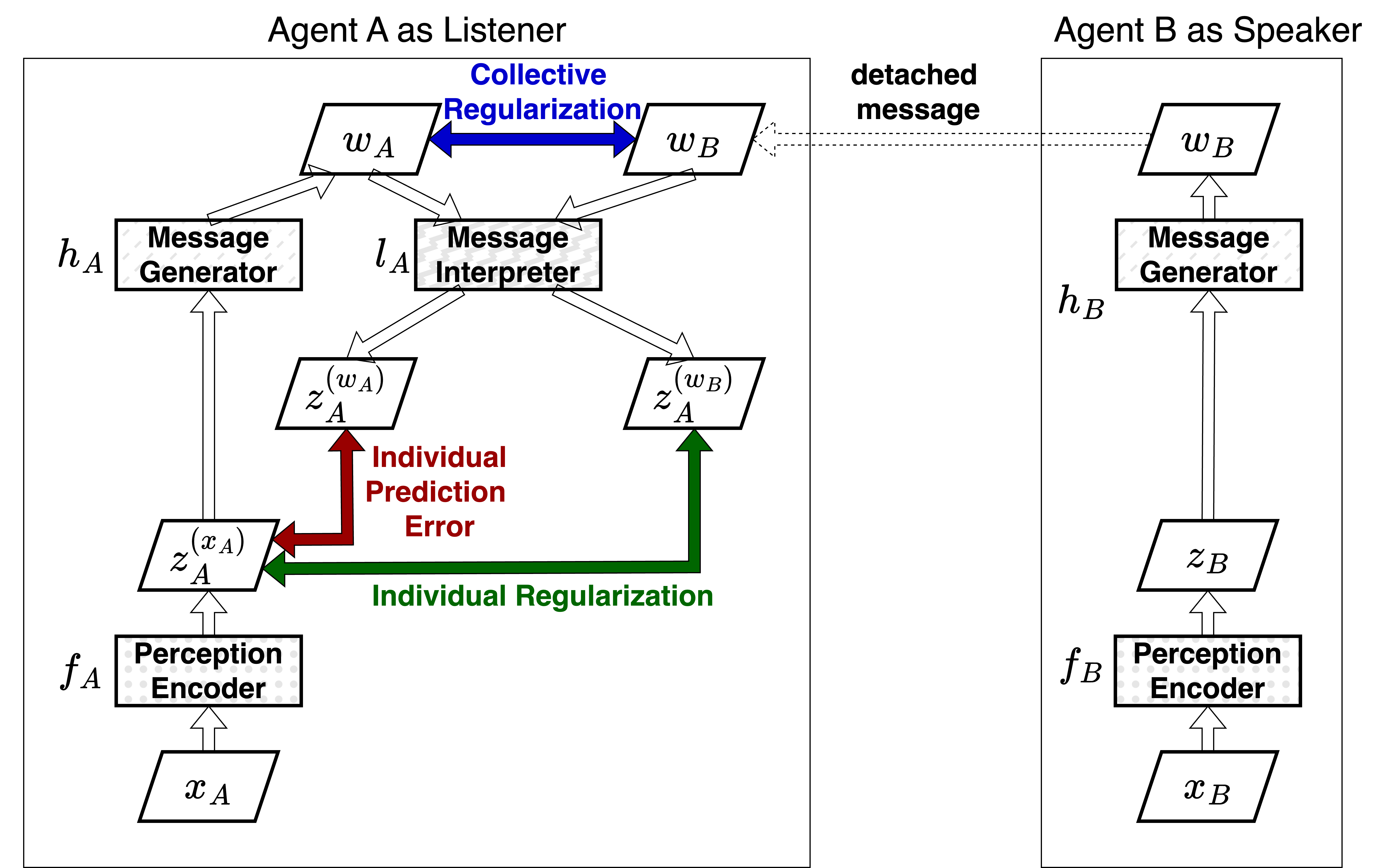}
\caption{Information flow in SSNG when Agent A acts as the listener and Agent B as the speaker. 
Agent B observes $x_B$, encodes it into a latent representation $z_B$, and produces a message $w_B$, which is transmitted to Agent A. 
Meanwhile, Agent A observes $x_A$, encodes it through its perception module $f_A$ into the perceptual latent $z_A^{(x_A)}$, and generates its own message $w_A$ using the message generator $h_A$. 
Both $w_A$ and the received $w_B$ are processed by Agent A’s message interpreter $l_A$ to produce reconstructed latents $z_A^{(w_A)}$ and $z_A^{(w_B)}$. 
The listener’s objective then compares the perceptual latent $z_A^{(x_A)}$ with these decoded latents to update its parameters.}
\label{fig: agent_arch}
\end{figure}


\begin{algorithm}[t]
\caption{SimSiam Naming Game}
\label{alg: SSNG}
\begin{algorithmic}[1]

\STATE \texttt{\# Perception encoders: $f_A(\cdot)$ and $f_B(\cdot)$}
\STATE \texttt{\# Message generators: $h_A(\cdot)$ and $h_B(\cdot)$}
\STATE \texttt{\# Message interpreters: $l_A(\cdot)$ and $l_B(\cdot)$}
\STATE \texttt{\# D: negative cosine similarity;\; CE: in-batch cross entropy}
\STATE \texttt{\# sg($\cdot$): stop-gradient operator}

\FOR{$t=1$ \TO \texttt{MaxIter}}
   \FOR{each mini-batch $\mathcal{B}=\{(x_{A}^{(i)},x_{B}^{(i)})\}_{i=1}^{|\mathcal{B}|}$}
      \STATE $z_A^{(x_A^{(i)})} \leftarrow f_A(x_A^{(i)}), \;\; z_B^{(x_B^{(i)})} \leftarrow f_B(x_B^{(i)})$ \texttt{\# Agents infer perceptual latents}
      \STATE $w_A^{(i)} \leftarrow h_A(z_A^{(x_A^{(i)})}), \;\; w_B^{(i)} \leftarrow h_B(z_B^{(x_B^{(i)})})$ \texttt{\# Agents generate messages}
      \STATE $z_A^{(w_A^{(i)})} \leftarrow l_A(w_A^{(i)}), \;\; z_B^{(w_B^{(i)})} \leftarrow l_B(w_B^{(i)})$ \texttt{\# Agents interpret own message}
      \STATE $z_A^{(w_B^{(i)})} \leftarrow l_A(w_B^{(i)}), \;\; z_B^{(w_A^{(i)})} \leftarrow l_B(w_A^{(i)})$ \texttt{\# Agents interpret partner's message}
      \STATE $\mathcal{W}_A \leftarrow \{w_A^{(i)}\}_{i=1}^{|\mathcal{B}|}, \quad \mathcal{W}_B \leftarrow \{w_B^{(i)}\}_{i=1}^{|\mathcal{B}|}$ \texttt{\# Messages in batch}
      \STATE $\mathcal{L}_A \leftarrow \sum_{i=1}^{|\mathcal{B}|} \Big[ \alpha D\!\left(z_A^{(x_A^{(i)})}, z_A^{(\texttt{sg}(w_B^{(i)}))}\right) + \beta D\!\left(z_A^{(x_A^{(i)})}, z_A^{(w_A^{(i)})}\right) + \gamma\, \mathrm{CE}_{\mathrm{in\text{-}batch}}\left(\texttt{sg}(w_B^{(i)}), \mathcal{W}_A\right) \Big]$
      \STATE $\mathcal{L}_B \leftarrow \sum_{i=1}^{|\mathcal{B}|} \Big[ \alpha D\!\left(z_B^{(x_B^{(i)})}, z_B^{(\texttt{sg}(w_A^{(i)}))}\right) + \beta D\!\left(z_B^{(x_B^{(i)})}, z_B^{(w_B^{(i)})}\right) + \gamma\, \mathrm{CE}_{\mathrm{in\text{-}batch}}\left(\texttt{sg}(w_A^{(i)}), \mathcal{W}_B\right) \Big]$
      \STATE $\mathcal{L} \leftarrow \mathcal{L}_A + \mathcal{L}_B$
      \STATE Update $f_A,h_A,l_A,f_B,h_B,l_B$ by minimizing $\mathcal{L}$ \texttt{\# Backpropagation}
   \ENDFOR
\ENDFOR

\end{algorithmic}
\end{algorithm}


\subsection{Communication via SSNG between two agents}
\label{subsec_ssng_emcom}
SSNG enables agents to iteratively align their internal representations and establish a shared EmLang through structured interactions while maintaining agent-specific perceptual encodings (Fig.~\ref{fig: SSNG_proc}). The inference process, involving both agents A and B, proceeds as follows:
\begin{align}
z_A &\sim q_{\phi}(z_A \mid x_A) \quad &\text{(Agent A infers $z_A$ from $x_A$)} \nonumber \\
z_B &\sim q_{\phi}(z_B \mid x_B) \quad &\text{(Agent B infers $z_B$ from $x_B$)} \nonumber \\
w &\sim q_{\phi}(w \mid z_A, z_B) \quad &\text{(The shared $w$ is inferred from $z_A$ and $z_B$)} \nonumber
\end{align}
The process unfolds as follows:
\begin{enumerate}
\item[\textbf{i)}] \textbf{Perception:} Both agents simultaneously process their respective inputs $x_A$ and $x_B$ through their perception encoders to generate internal representations $z_A = f_A(x_A)$ and $z_B = f_B(x_B)$
\item[\textbf{ii)}] \textbf{Naming:} Each agent generates a message from its latent representation $w_A = h_A(z_A)$ and $w_B = h_B(z_B)$. The messages are then exchanged between the agents.
\item[\textbf{iii)}] \textbf{Interpretation:} Each agent interprets the partner's received message into a latent reconstruction $z_A^{(w_B)} = l_A(w_B)$ and $z_B^{(w_A)} = l_B(w_A)$. Each agent also decodes its own message, producing $z_A^{(w_A)} = l_A(w_A)$ and $z_B^{(w_B)} = l_B(w_B)$
\item[\textbf{iv)}] \textbf{Learning:} Each agent simultaneously computes its listener objective (Eq.~\ref{eq: loss_ssng_agent}), comparing its perceptual latent against  
(a) the latent decoded from the partner's message and 
(b) the latent decoded from its generated message. Parameters are updated in parallel.
\item[\textbf{v)}] \textbf{Iteration:} The above steps are repeated until the maximum number of iterations is reached.
\end{enumerate}

Each agent aims to align its predicted message $w_{i}$ with the received message $w_{j}$ from its partner by maximizing their inner product $w_{i}^\top w_{j}$. However, directly optimizing this pairwise alignment objective is prone to representational collapse \citep{2021_chen_simsiam}. We therefore optimize a cross-entropy \emph{surrogate loss} that approximates the intended inner-product alignment while providing stable gradients. Specifically, for each received $w_{j}$, the corresponding prediction $w_{i}$ is treated as a positive pair, while predictions $\{w'_{i}\}$ from other samples in the mini-batch serve as negative pairs. Importantly, the received message $w_{j}$ is treated as fixed via a stop-gradient operation, ensuring that no gradients propagate to the partner agent and preserving the fully feedback-free nature of SSNG. The minimization form replaces cosine-similarity maximization with negative cosine similarity and uses in-batch cross-entropy as the surrogate for message alignment, as detailed in Algorithm~\ref{alg: SSNG}. 
 
This iterative communication, grounded in the CPC hypothesis \citep{2024_taniguchi_CPC, 2025_taniguchi_cpc_science}, allows agents to update their understanding by encoding, sharing, decoding, and learning from shared symbols. A comparison of SSNG with MHNG~\citep{2023_taniguchi_inter_gmm_vae}, referential game~\citep{2017_lazaridou_referential}, and reconstruction game~\citep{2020_chaabouni_recon_compositionality_generalization} is provided in Table~\ref{tab: EmCom_game_comparison} (\ref{appendix_emcom_comparison}).


\subsection{SSNG for Representation Learning}\label{sec: ssng_ssl}
Although SSNG was originally introduced as a two-agent framework for EmCom, its mechanism of aligning internal representations through message exchange naturally reduces to SSL. When the two agents are reinterpreted as two augmented views processed within a single system, SSNG reduces to a SimSiam-like architecture. In this setting, the intermediate variable $w$ functions as a stochastic bottleneck between the projected representation and the predicted representation. When the distribution over $w$ collapses to a point mass, this stochastic bottleneck becomes deterministic, recovering the predictor structure used in SimSiam \citep{2021_chen_simsiam}.

Under this interpretation, the perception module $f$ (backbone + projector) maps an input image $x$ to a projected representation $z$. The generator–interpreter pair $(h,l)$ together form a predictor $g = l \circ h$, which maps $z$ to a predicted representation $\hat z$. Given two augmented views $x_A$ and $x_B$ of the same image, the perception module produces
\[
z_A = f(x_A), \qquad z_B = f(x_B).
\]
During training, one branch serves as the prediction branch, where $z_A$ is passed through the predictor $g$ to produce $\hat z_A = g(z_A)$. The other branch serves as the target branch, where $z_B$ is treated as a stop-gradient target as in SimSiam. The predicted representation $\hat z_A$ is then aligned with $z_B$. Swapping the roles of $x_A$ and $x_B$ yields the complementary term, forming a symmetric alignment objective. The corresponding graphical model is illustrated in Fig.~\ref{fig: SSNG_SSL}(a).

\paragraph{Variational formulation}
The SSL interpretation of SSNG can also be derived from the same variational framework used in the two-agent case (Section~\ref{subsec_2agent}). In the representation learning setting, we consider a pair of augmented views $\mathbb{X}=\{x_A,x_B\}$. The corresponding latent variables $\mathbb{Z}=\{z_A,z_B\}$ are treated as two stochastic views of a shared latent representation $z$. The model parameters $\theta$ and $\phi$ are optimized by maximizing the ELBO:
\begin{equation}
\theta^*, \phi^* =
\arg\max_{\theta,\phi}
\mathbb{E}_{q_{\phi}(w,z|\mathbb{X})}
\left[
\log
\frac{p_{\theta}(w,z,\mathbb{X})}
     {q_{\phi}(w,z|\mathbb{X})}
\right],
\end{equation}

Following the variational interpretation of SSL \citep{2023_nakamura_visimsiam}, both the sign variable $w$ and the projected representation $z$ are modeled as directional variables on hyperspheres with uniform priors. The variational distribution over $w$ is modeled using the Power Spherical (PS) distribution \citep{2020_decao_powerspherical}. 
Unlike the vMF distribution, the PS distribution admits a simple reparameterization that enables gradient-based optimization, which is required in the SSL setting where $w$ can be sampled differentiably. The latent representation $z$ is inferred from both augmented views through an MoE formulation. Throughout this section, we denote the two branches by indices $i,j \in \{A,B\}$ with $i \neq j$, where $i$ denotes the predicted branch and $j$ denotes the target branch. With $M=2$ corresponding to the two augmented views, the distributions are defined as follows:
\begin{align}
p(w) &:= \mathcal{U}(S^{d_w-1}) 
\label{eq: ssl_w_prior} \\
p(z) &:= \mathcal{U}(S^{d_z-1}) 
\label{eq: ssl_z_prior} \\
q_{\phi}(w \mid z) &:= \text{PS}(w; \mu_w = h_{(\mu)}(z), \kappa_w = h_{(\kappa)}(z)) 
\label{eq: ssl_w_variational} \\
q_{\phi}(z \mid \mathbb{X}) &:= \frac{1}{M} \sum_{i=1}^{M} \text{vMF}(z; \mu_i^{(q_z)} = f(x_i), \kappa^{(q_z)}) 
\label{eq: ssl_z_variational} \\
p_\theta(z \mid \mathbb{X}) &:= \prod_{i=1}^{M} \text{vMF}(z; \mu_i^{(p_z)} = g(x_j), \kappa^{(p_z)})
\label{eq: ssl_z_posterior}
\end{align}
where 
\begin{itemize}
\item $h_{(\mu)}$ and $h_{(\kappa)}$ produce the mean direction and concentration parameter of the PS distribution, respectively.
\item $\hat z_j = g(x_j) = l \big( h( f(x_j) ) \big)$ denotes the prediction of one augmented view, while $z_i = f(x_i)$ denotes the projected representation of the other view.
\end{itemize}

\paragraph{Learning objective}
In the SSL setting, the two branches correspond to the augmented views $A$ and $B$. Thus, the pair $(i,j)$ represents either $(A,B)$ or $(B,A)$. The resulting learning objective consists of a representation alignment term and a regularization term:
\begin{align}
\mathcal{J}_{\mathrm{SSL}} &:= \mathcal{J}_{\text{align}} - \beta_{\mathrm{KL}} \, \mathcal{J}_{\mathrm{KL}} 
\end{align}
where
\begin{align}
\mathcal{J}_{\mathrm{align}}
&:= (\hat z_A)^{\top}(z_B) + (\hat z_B)^{\top}(z_A), \\
\mathcal{J}_{\mathrm{KL}} 
&:= D_{\mathrm{KL}} \Big( q_{\phi}(w \mid z_A) \;\big\|\; p(w) \Big) + D_{\mathrm{KL}} \Big( q_{\phi}(w \mid z_B) \;\big\|\; p(w) \Big)
\end{align}
Here $D_{\mathrm{KL}}$ denotes the KL divergence. Under the PS distribution, these KL divergences admit closed-form expressions and can therefore be computed directly \citep{2020_decao_powerspherical}. The parameter $\beta_{\mathrm{KL}}$ controls the weight of the regularization term. The derivation is provided in~\ref{appendix_ssng_elbo_representation}. These terms can be interpreted as follows:
\begin{itemize}

\item \textbf{Alignment terms $\mathcal{J}_{\mathrm{align}}$.}
In SSNG, two agents align their latent representations through message exchange, producing the \emph{Individual Prediction Error} and \emph{Collective Regularization} terms. 
In the SSL interpretation, the two agents are replaced by two augmented views processed within a single system, and these two components collapse into the symmetric SimSiam-style alignment objective.

\item \textbf{Regularization term $\mathcal{J}_{\mathrm{KL}}$.}
This term regularizes the stochastic bottleneck $w$ by maximizing the entropy of the PS distribution. In SSNG, the prior over $w$ is induced by the partner’s message, whereas in the SSL formulation $w$ follows a uniform hyperspherical prior, acting as a regularized stochastic bottleneck within the single-system architecture.

\end{itemize}

\paragraph{Stop-gradient mechanism}
To prevent representational collapse, SSNG adopts the stop-gradient mechanism introduced in SimSiam~\citep{2021_chen_simsiam}. During alignment, one branch is treated as a fixed target while the other branch predicts its representation via the predictor $g=l\circ h$. This asymmetric gradient flow stabilizes training and discourages trivial solutions. Empirical comparisons with and without the stop-gradient mechanism are reported in Section~\ref{subsec_exp_ssl}. The training procedure is summarized in Algorithm~\ref{alg: SSNG-repLrn}.

In summary, SSNG can be interpreted as a communication-based representation learning framework, while the SSL formulation corresponds to the single-system limit in which agents are replaced by augmented views.


\begin{algorithm}[t]
\caption{SSNG in Representation Learning}
\label{alg: SSNG-repLrn}
\begin{algorithmic}[1]
\STATE \texttt{\# Backbone + projector: $f(\cdot)$;\; generator: $h(\cdot)$;\; interpreter: $l(\cdot)$}
\STATE \texttt{\# $D$: negative cosine similarity;\; KL: KL divergence}
\STATE \texttt{\# sg($\cdot$): stop-gradient operator}

\FOR{each mini-batch $x \in \mathcal{B}$}
    \STATE $x_A \leftarrow \mathrm{aug}(x), \quad x_B \leftarrow \mathrm{aug}(x)$ \texttt{\# Two augmented views}
    \STATE $z_A \leftarrow f(x_A), \quad z_B \leftarrow f(x_B)$ \texttt{\# Infer projected representations}
    \STATE $(\mu_A, \kappa_A) \leftarrow h(z_A), \quad (\mu_B, \kappa_B) \leftarrow h(z_B)$ \texttt{\# Produce PS parameters}
    \STATE $w_A \sim \mathrm{PS}(\mu_A, \kappa_A), \quad w_B \sim \mathrm{PS}(\mu_B, \kappa_B)$ \texttt{\# Reparameterized samples}
    \STATE $\hat{z}_A \leftarrow l(w_A), \quad \hat{z}_B \leftarrow l(w_B)$ \texttt{\# Infer predicted representations}
    \STATE $\mathcal{L}_{\mathrm{align}} \leftarrow D(\hat{z}_A, \texttt{sg}(z_B)) + D(\hat{z}_B, \texttt{sg}(z_A))$
    \STATE $\mathcal{L}_{\mathrm{KL}} \leftarrow D_{\mathrm{KL}}\!\left(q_{\phi}(w\mid z_A)\|p(w)\right) + D_{\mathrm{KL}}\!\left(q_{\phi}(w\mid z_B)\|p(w)\right)$
    \STATE $\mathcal{L} \leftarrow \mathcal{L}_{\mathrm{align}} - \beta_{\mathrm{KL}}\,\mathcal{L}_{\mathrm{KL}}$
    \STATE Update $f,h,l$ by minimizing $\mathcal{L}$ \texttt{\# Backpropagation}
\ENDFOR

\end{algorithmic}
\end{algorithm}


%
%

\section{Experiments}
We conduct two sets of experiments to evaluate the two main claims of this paper. First, we investigate whether SSNG enables agents to develop informative symbolic communication through interaction. Second, we examine whether SSNG recovers competitive SSL performance when the two agents are interpreted as two augmented views within a single system. In both cases, representations are evaluated using a linear classifier with top-1 accuracy.
\begin{itemize}
\item \textbf{Experiment 1 (EmCom).} We evaluate whether the discrete messages emerging through two-agent interaction retain class-relevant semantic information, assessed via linear-probe classification on the learned message representations.
\item \textbf{Experiment 2 (SSL).} We evaluate SSNG under the SSL interpretation, where the two agents are merged into a single network processing two augmented views, and report linear evaluation performance following standard SSL protocols \citep{2020_chen_simclr}.
\end{itemize}
Our implementation and training code\footnote{https://github.com/nlhoang/SimSiamNamingGame} are publicly available.


\subsection{Experiment 1: SSNG in EmCom}  
\paragraph{Description} This experiment evaluates whether the discrete messages $w$ generated by SSNG retain class-relevant information from visual inputs. We measure this using top-1 accuracy from a linear classifier trained on frozen messages. The core question is: Do SSNG agents develop discrete messages that remain discriminative with respect to object categories?

\paragraph{Datasets} We conduct experiments on CIFAR-10~\citep{2009_krizhevsky_cifar} and on a 20-class subset of ImageNet-100~\citep{2009_deng_imagenet} (e.g., dog, cat, airplane, ship, train), covering animals, vehicles, household objects, etc.

\paragraph{Heterogeneous Agents} To test robustness, the two agents use different perceptual backbones:
ResNet-18 and ResNet-34~\citep{2016_he_resnet}, each paired with its own language coder. All backbones are initialized with ImageNet weights and fully trainable, ensuring that any shared communication system emerges from interaction rather than architectural symmetry.

\paragraph{Baselines} We compare SSNG to four established EmCom baselines:
\begin{itemize}
\item No Communication (NoCom): Agents receive the same training augmentations and objectives but never exchange messages. 
\item Referential Game: A classical supervised signal-passing framework where the speaker produces a message and the listener selects a target from among multiple candidates, including distractors.
\item Reconstruction Game: The listener reconstructs the speaker’s perceptual embedding, providing direct feedback via reconstruction loss.
\item Metropolis–Hastings Naming Game (MHNG): A feedback-free communication system where messages and categories emerge through the Metropolis–Hastings algorithm.
\end{itemize}
All baselines use the same data augmentations, the same perceptual backbones with pretrained initialization, with all layers unfrozen. 

\paragraph{Protocol} Each image is shown to both agents with independent augmentations. SSNG, referential, and reconstruction models were trained for 50 epochs; MHNG required 200 epochs due to slower sampling convergence. Message length was fixed to 10 tokens to enforce a strict communication bottleneck. Vocabulary size was set to 100 for CIFAR-10 and 1,000 for the more semantically complex ImageNet-100 dataset.

Discrete messages are produced via the Gumbel–Softmax relaxation \citep{2017_jang_categorical, 2017_maddison_categorical} with a Straight-Through estimator \citep{2013_bengio_ste}. To stabilize early optimization and prevent premature representation collapse, we apply temperature annealing and an initial warm-up phase using continuous (soft) tokens before transitioning to fully discrete messages. Additional architectural and optimization details are  in~\ref{appendix_experiment_settings}.

\paragraph{Evaluation}
After training, agent parameters were frozen to extract discrete messages. To evaluate semantic richness, we trained a linear classifier on the training messages (formatted as flattened one-hot vectors) and reported top-1 accuracy on the test set.

Furthermore, unlike one-sided referential and reconstruction games, SSNG’s symmetric architecture enables bidirectional communication, allowing us to uniquely quantify the convergence of a shared language. For SSNG, MHNG, and a NoCom baseline, we measured cross-agent alignment using cosine similarity (for message embeddings) and token-level overlap (for discrete symbols) between messages produced for the same image. Higher values indicate better performance for all metrics.


\begin{table}[t]
\centering
\caption{Evaluation of discrete messages on CIFAR-10, including message Top-1 classification accuracy and inter-agent message agreement measured via cosine similarity and overlap coefficient. Highest values are in bold, and second-highest values are underlined.}
\label{tab:exp1_cifar10}
\begin{tabular}{l l c c c}
\toprule
\textbf{Setting} & \textbf{Agent} & \textbf{Top1 Acc} & \textbf{CosSim} & \textbf{Overlap} \\
\midrule
Referential Game   		& Speaker 	& 23.17  {\scriptsize$\pm$2.46} & - & -  \\
\midrule
Reconstruction Game   	& Speaker 	& \underline{38.94}  {\scriptsize$\pm$3.79} & - & -  \\
\midrule
NoCom   				& Agent A 		& 11.29 {\scriptsize$\pm$3.89} & 0.04 {\scriptsize$\pm$0.05} & 0.09 {\scriptsize$\pm$0.05} \\
        					& Agent B 	& 12.67 {\scriptsize$\pm$4.02} &  &  \\
\midrule
MHNG   				& Agent A 		& 34.52 {\scriptsize$\pm$3.54} & 0.27 {\scriptsize$\pm$0.11} & 0.28 {\scriptsize$\pm$0.08} \\
        					& Agent B 	& 32.26 {\scriptsize$\pm$3.73} &  &    \\
\midrule
\textbf{SSNG}   		& Agent A 		& \textbf{59.31} {\scriptsize$\pm$2.72} & \textbf{0.83} {\scriptsize$\pm$0.09} & \textbf{0.84} {\scriptsize$\pm$0.05} \\
        					& Agent B 	& \textbf{59.36} {\scriptsize$\pm$2.83} &  &    \\
\bottomrule
\end{tabular}
\end{table}


\begin{table}[t]
\centering
\caption{Evaluation of discrete messages on 20-class subset of ImageNet100, including message Top-1 classification accuracy and inter-agent message agreement measured via cosine similarity and overlap coefficient. Highest values are in bold, and second-highest values are underlined.}
\label{tab:exp1_imagenet100}
\begin{tabular}{l l c c c}
\toprule
\textbf{Setting} & \textbf{Agent} & \textbf{Top1 Acc} & \textbf{CosSim} & \textbf{Overlap} \\
\midrule
Referential Game   		& Speaker 	& \underline{27.01} {\scriptsize$\pm$1.59} & - & -  \\
\midrule
Reconstruction Game   	& Speaker 	& 25.95 {\scriptsize$\pm$3.69} & - & -  \\
\midrule
NoCom   				& Agent A 		& 6.32 {\scriptsize$\pm$1.17} &	$(5.0 \pm 0.8)\times10^{-4}$ 
					&$(6.0 \pm 1.1)\times10^{-3}$ \\
        					& Agent B 	& 6.92 {\scriptsize$\pm$1.59} &  &  \\
\midrule
MHNG   				& Agent A 		& 18.72 {\scriptsize$\pm$1.77} & 0.09 {\scriptsize$\pm$0.03
} & 0.11 {\scriptsize$\pm$0.03} \\
        					& Agent B 	& 19.80 {\scriptsize$\pm$1.75} &  &    \\
\midrule
\textbf{SSNG}    		& Agent A 		& \textbf{41.98} {\scriptsize$\pm$3.78} & \textbf{0.21} {\scriptsize$\pm$0.08
} & \textbf{0.22} {\scriptsize$\pm$0.09} \\
        					& Agent B 	& \textbf{42.83} {\scriptsize$\pm$3.24} &  &    \\
\bottomrule
\end{tabular}
\end{table}


\paragraph{Results and Discussion}
All experiments were repeated five times with different random seeds, and we report the mean results in Table~\ref{tab:exp1_cifar10} and Table~\ref{tab:exp1_imagenet100}. The main observations are as follows.
\begin{itemize}

\item On both CIFAR-10 and the ImageNet-100 subset, SSNG achieves the highest top-1 accuracy, reaching $51.12\%$ and $51.06\%$ on CIFAR-10 and $44.60\%$ and $42.91\%$ on ImageNet-100 for Agents A and B respectively. These results substantially outperform MHNG, the reconstruction game, the referential game, and the NoCom setting, indicating that messages learned through SSNG preserve more class-relevant information than those produced by other communication frameworks.

\item As the dataset complexity increases from CIFAR-10 to ImageNet-100, the performance of all methods decreases. Nevertheless, SSNG maintains a large margin over the baselines on both datasets, suggesting that the learned messages remain informative even under more complex visual conditions.

\item Cross-agent message alignment shows a similar trend. SSNG achieves substantially higher cosine similarity and overlap coefficients than MHNG and NoCom. In the NoCom condition, agents never exchange messages, resulting in minimal similarity. MHNG also shows reduced similarity on ImageNet-100, likely due to higher rejection rates in the Metropolis–Hastings updates, even with four times as many epochs (200 vs.\ 50). These results indicate that SSNG enables agents to develop a more consistent shared communication protocol while preserving class-discriminative information in the messages.

\item The use of heterogeneous agents with different perceptual backbones highlights the robustness of SSNG. Despite architectural differences, the agents are still able to develop a consistent communication system, supporting the CPC perspective that EmCom arises through representation alignment rather than architectural symmetry.
\end{itemize}

\begin{figure}[!htbp]
  \centering
  \begin{tabular}{cc}
    \includegraphics[width=0.42\linewidth]{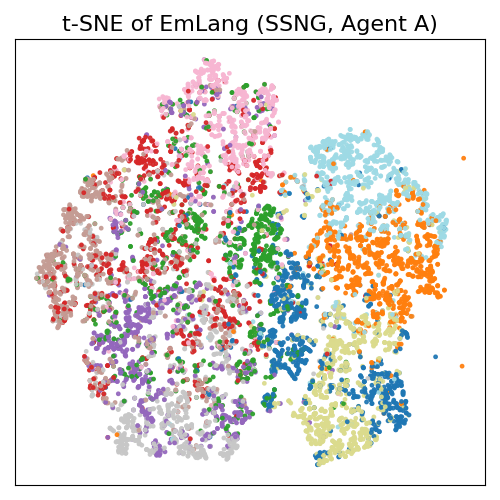} &
    \includegraphics[width=0.42\linewidth]{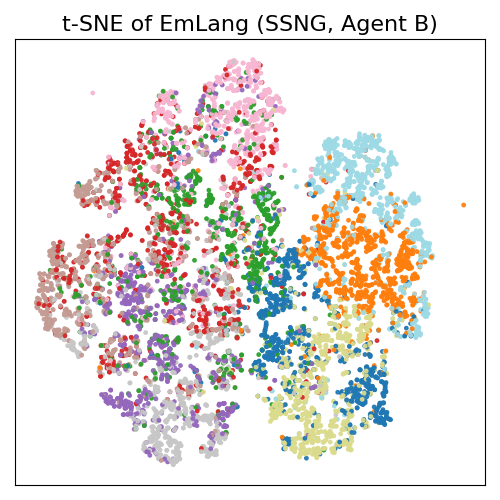} \\
    \includegraphics[width=0.42\linewidth]{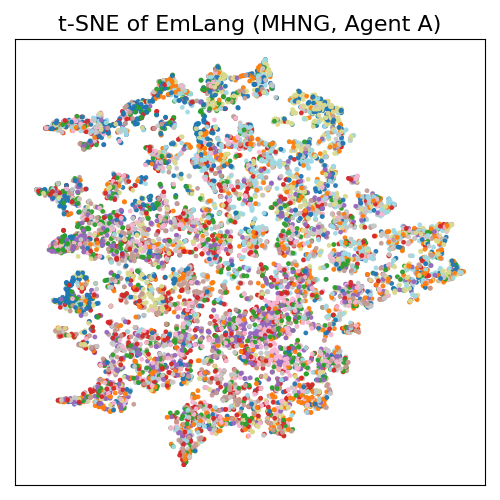} &
    \includegraphics[width=0.42\linewidth]{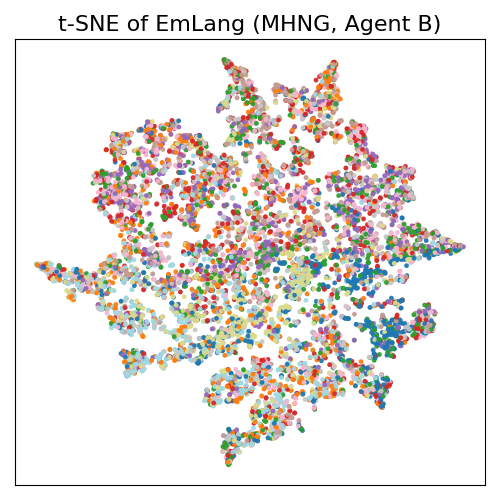} \\
    \includegraphics[width=0.42\linewidth]{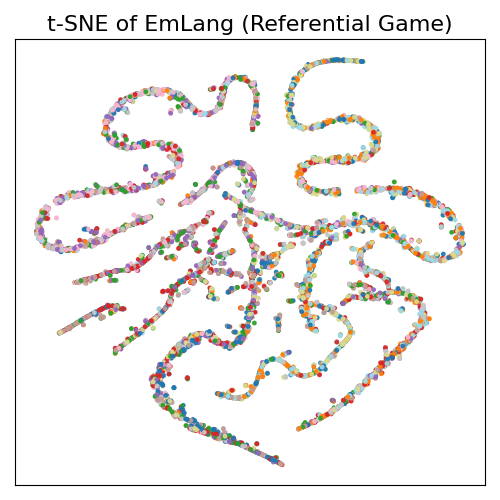} &
    \includegraphics[width=0.42\linewidth]{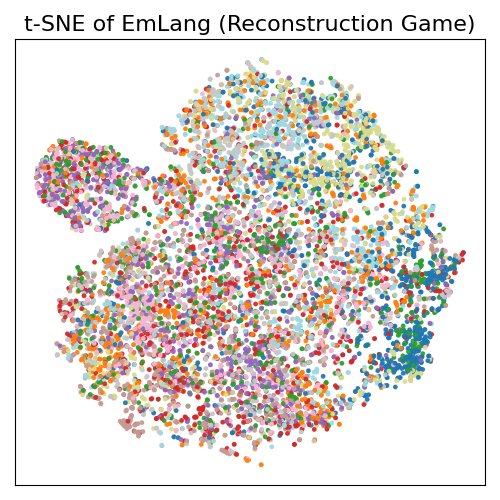} \\
  \end{tabular}
  \caption{{t-SNE visualization of emergent message representations on CIFAR-10 (validation set).}
Points correspond to images and are colored by class label. The top row shows SSNG with two agents, the middle row shows MHNG with two agents, and the bottom row shows the referential and reconstruction games. Message representations are computed from discrete message sequences, and the same t-SNE settings are used across all methods.}
  \label{fig:tsne_messages_cifar10}
\end{figure}

\begin{figure}[!htbp]
\centering
\begin{tabular}{cccc}

\begin{minipage}[t]{0.23\textwidth}
\centering
\includegraphics[width=.9\textwidth]{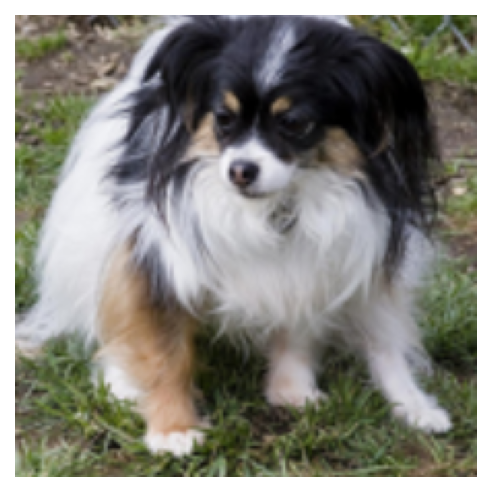} \\
{\scriptsize
\textbf{A:} [\textcolor{classagree}{692} * \textcolor{classagree}{315} 507 ***** \textcolor{agentagree}{710}]\\
\textbf{B:} [\textcolor{classagree}{686} * \textcolor{classagree}{943} \textcolor{classagree}{378} ***** \textcolor{agentagree}{710}]
}
\end{minipage}
&
\begin{minipage}[t]{0.23\textwidth}
\centering
\includegraphics[width=.9\textwidth]{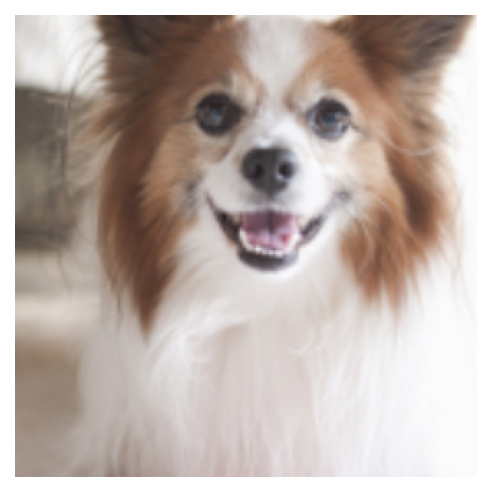} \\
{\scriptsize
\textbf{A:} [\textcolor{classagree}{692} * \textcolor{classagree}{315} \textcolor{classagree}{807} ***** \textcolor{agentagree}{710}]\\
\textbf{B:} [868 * \textcolor{classagree}{943} \textcolor{classagree}{378} ***** 706]
}
\end{minipage}
&
\begin{minipage}[t]{0.23\textwidth}
\centering
\includegraphics[width=.9\textwidth]{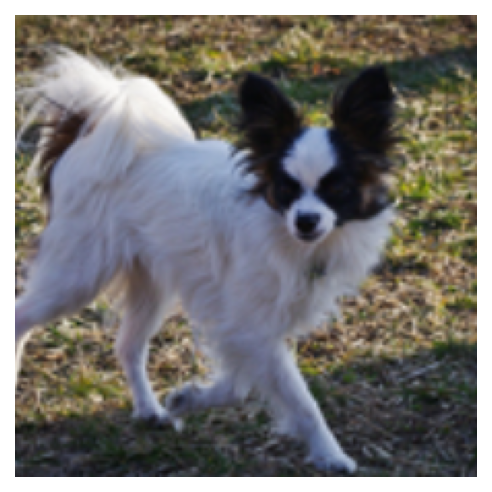} \\
{\scriptsize
\textbf{A:} [\textcolor{classagree}{692} * \textcolor{classagree}{315} \textcolor{classagree}{807} ***** \textcolor{agentagree}{710}]\\
\textbf{B:} [\textcolor{classagree}{686} * \textcolor{classagree}{943} \textcolor{classagree}{378} ***** \textcolor{agentagree}{710}]
}
\end{minipage}
&
\begin{minipage}[t]{0.23\textwidth}
\centering
\includegraphics[width=.9\textwidth]{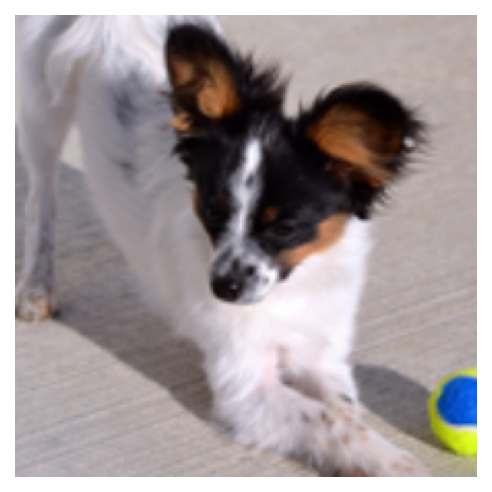} \\
{\scriptsize
\textbf{A:} [\textcolor{classagree}{692} * \textcolor{classagree}{315} \textcolor{classagree}{807} ***** \textcolor{agentagree}{710}]\\
\textbf{B:} [\textcolor{classagree}{686} * \textcolor{classagree}{943} \textcolor{classagree}{378} ***** 706]
}
\end{minipage}
\\

\begin{minipage}[t]{0.23\textwidth}
\centering
\includegraphics[width=.9\textwidth]{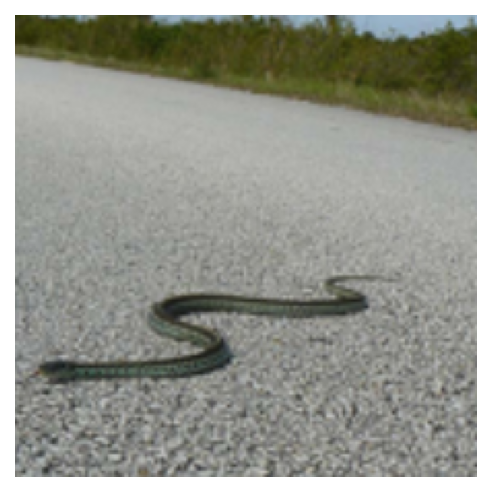} \\
{\scriptsize
\textbf{A:} [\textcolor{agentagree}{17} *** \textcolor{agentagree}{248} \textcolor{agentagree}{258} ** \textcolor{agentagree}{878} *]\\
\textbf{B:} [\textcolor{agentagree}{17} *** \textcolor{agentagree}{248} \textcolor{agentagree}{258} ** \textcolor{agentagree}{878} *]
}
\end{minipage}
&
\begin{minipage}[t]{0.23\textwidth}
\centering
\includegraphics[width=.9\textwidth]{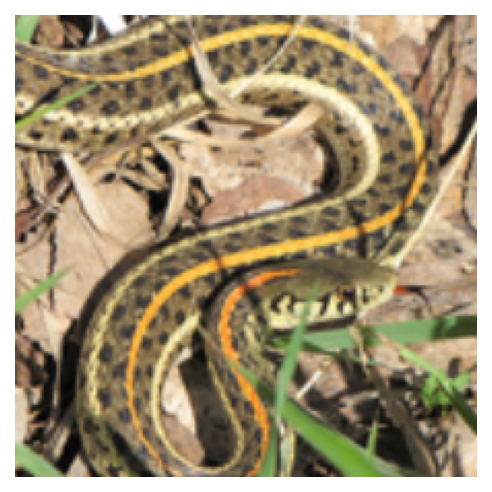} \\
{\scriptsize
\textbf{A:} [\textcolor{classagree}{330} \textcolor{classagree}{704} ** \textcolor{agentagree}{248} \textcolor{agentagree}{258} ****]\\
\textbf{B:} [\textcolor{classagree}{1} \textcolor{classagree}{353} ** \textcolor{agentagree}{248} \textcolor{agentagree}{258} ****]
}
\end{minipage}
&
\begin{minipage}[t]{0.23\textwidth}
\centering
\includegraphics[width=.9\textwidth]{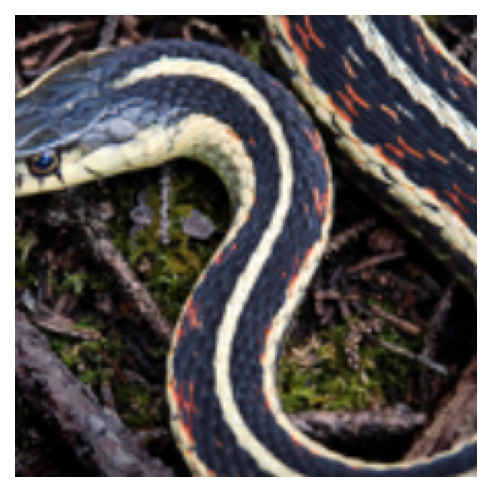} \\
{\scriptsize
\textbf{A:} [\textcolor{classagree}{330} \textcolor{classagree}{704} ** \textcolor{agentagree}{248} \textcolor{agentagree}{258} ****]\\
\textbf{B:} [\textcolor{classagree}{1} *** \textcolor{agentagree}{248} \textcolor{agentagree}{258} \textcolor{agentagree}{878} ***]
}
\end{minipage}
&
\begin{minipage}[t]{0.23\textwidth}
\centering
\includegraphics[width=.9\textwidth]{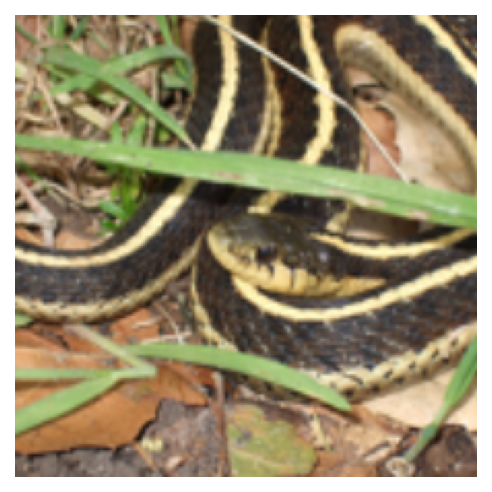} \\
{\scriptsize
\textbf{A:} [\textcolor{classagree}{330} \textcolor{classagree}{704} ** \textcolor{agentagree}{248} \textcolor{agentagree}{258} ****]\\
\textbf{B:} [\textcolor{classagree}{1} \textcolor{classagree}{353} ** \textcolor{agentagree}{248} \textcolor{agentagree}{258} ****]
}
\end{minipage}
\\ 

\begin{minipage}[t]{0.23\textwidth}
\centering
\includegraphics[width=.9\textwidth]{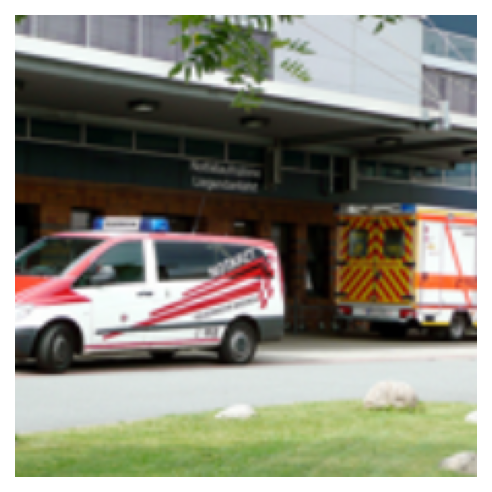} \\
{\scriptsize
\textbf{A:} [\textcolor{agentagree}{14} \textcolor{classagree}{926} \textcolor{agentagree}{956} ****** 347]\\
\textbf{B:} [\textcolor{agentagree}{14} 387 \textcolor{agentagree}{956} ****** 987]
}
\end{minipage}
&
\begin{minipage}[t]{0.23\textwidth}
\centering
\includegraphics[width=.9\textwidth]{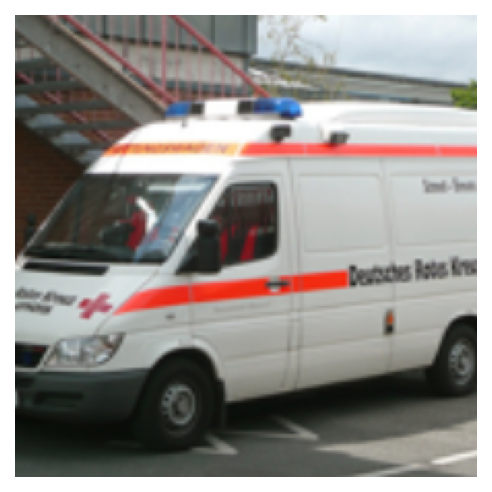} \\
{\scriptsize
\textbf{A:} [\textcolor{agentagree}{14} \textcolor{classagree}{926} * \textcolor{classagree}{83} ***** \textcolor{classagree}{276}]\\
\textbf{B:} [\textcolor{agentagree}{14} \textcolor{classagree}{309} * \textcolor{classagree}{459} ***** \textcolor{classagree}{651}]
}
\end{minipage}
&
\begin{minipage}[t]{0.23\textwidth}
\centering
\includegraphics[width=.9\textwidth]{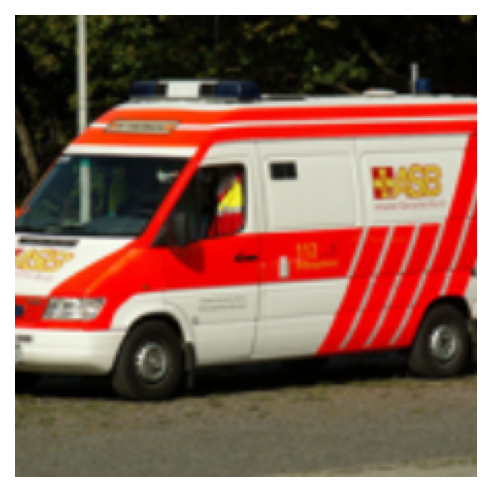} \\
{\scriptsize
\textbf{A:} [\textcolor{agentagree}{14} \textcolor{classagree}{926} * \textcolor{classagree}{83} ***** \textcolor{classagree}{276}]\\
\textbf{B:} [\textcolor{agentagree}{14} \textcolor{classagree}{309} * \textcolor{classagree}{459} ***** \textcolor{classagree}{651}]
}
\end{minipage}
&
\begin{minipage}[t]{0.23\textwidth}
\centering
\includegraphics[width=.9\textwidth]{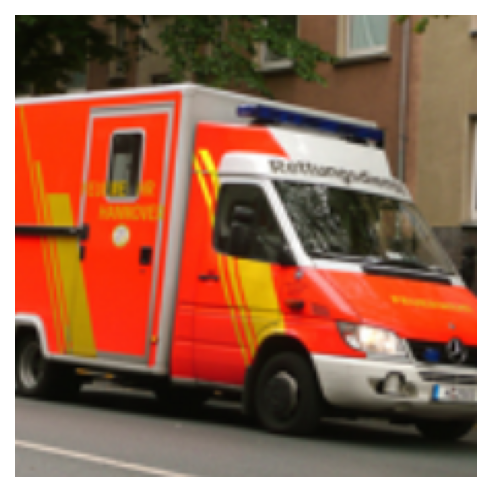} \\
{\scriptsize
\textbf{A:} [\textcolor{agentagree}{14} \textcolor{classagree}{926} * \textcolor{classagree}{83} ***** \textcolor{classagree}{276}]\\
\textbf{B:} [\textcolor{agentagree}{14} \textcolor{classagree}{309} * \textcolor{classagree}{459} ***** \textcolor{classagree}{651}]
}
\end{minipage}
\\ 

\begin{minipage}[t]{0.23\textwidth}
\centering
\includegraphics[width=.9\textwidth]{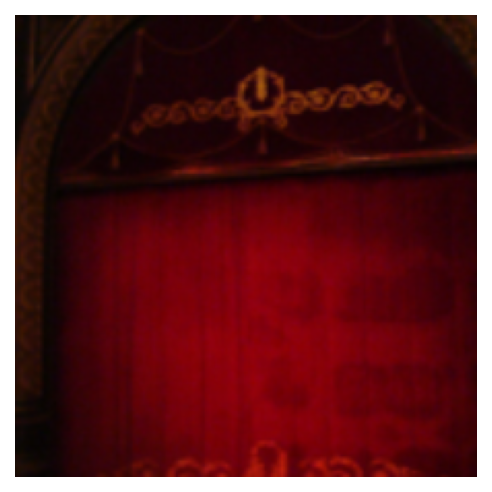} \\
{\scriptsize
\textbf{A:} [\textcolor{agentagree}{14} * \textcolor{agentagree}{956} * \textcolor{agentagree}{980} **** \textcolor{classagree}{718}]\\
\textbf{B:} [\textcolor{agentagree}{14} * \textcolor{agentagree}{956} * \textcolor{agentagree}{980} **** \textcolor{classagree}{651}]
}
\end{minipage}
&
\begin{minipage}[t]{0.23\textwidth}
\centering
\includegraphics[width=.9\textwidth]{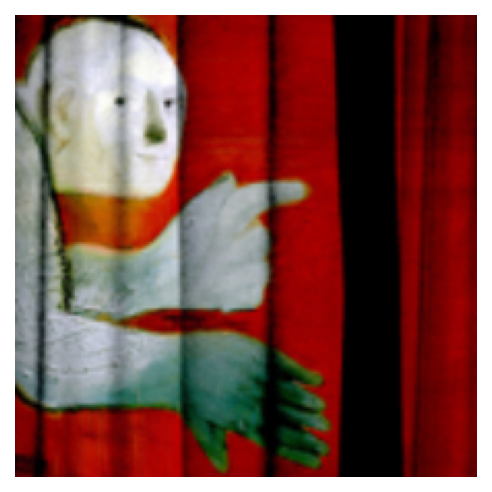} \\
{\scriptsize
\textbf{A:} [\textcolor{agentagree}{14} * \textcolor{agentagree}{956} * \textcolor{agentagree}{980} **** \textcolor{classagree}{718}]\\
\textbf{B:} [\textcolor{agentagree}{14} * \textcolor{agentagree}{956} * \textcolor{agentagree}{980} **** \textcolor{classagree}{651}]
}
\end{minipage}
&
\begin{minipage}[t]{0.23\textwidth}
\centering
\includegraphics[width=.9\textwidth]{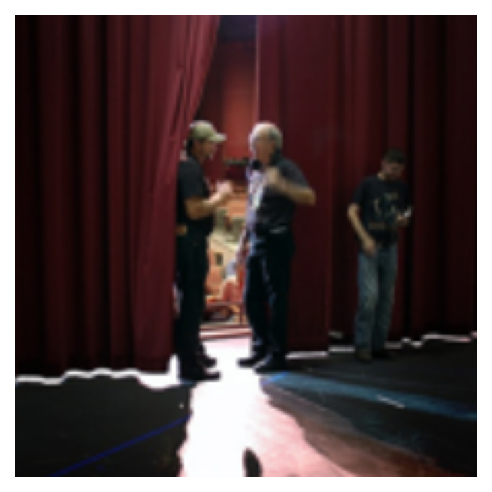} \\
{\scriptsize
\textbf{A:} [\textcolor{agentagree}{14} * \textcolor{agentagree}{956} * \textcolor{agentagree}{980} **** \textcolor{classagree}{718}]\\
\textbf{B:} [\textcolor{agentagree}{14} * \textcolor{agentagree}{956} * \textcolor{agentagree}{980} **** \textcolor{classagree}{651}]
}
\end{minipage}
&
\begin{minipage}[t]{0.23\textwidth}
\centering
\includegraphics[width=.9\textwidth]{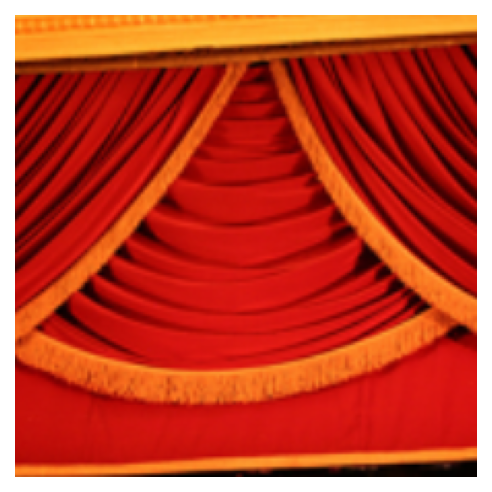} \\
{\scriptsize
\textbf{A:} [\textcolor{agentagree}{14} *** \textcolor{agentagree}{980} *** \textcolor{agentagree}{878} \textcolor{classagree}{718}]\\
\textbf{B:} [\textcolor{agentagree}{14} *** \textcolor{agentagree}{980} *** \textcolor{agentagree}{878} \textcolor{classagree}{651}]
}
\end{minipage}
\\ 

\begin{minipage}[t]{0.23\textwidth}
\centering
\includegraphics[width=.9\textwidth]{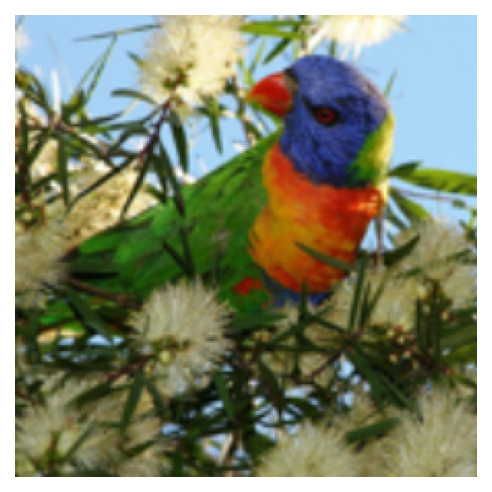} \\
{\scriptsize
\textbf{A:} [** \textcolor{agentagree}{956} \textcolor{classagree}{715} * \textcolor{agentagree}{258} \textcolor{classagree}{834} ***]\\
\textbf{B:} [** \textcolor{agentagree}{956} \textcolor{classagree}{777} * \textcolor{agentagree}{258} \textcolor{classagree}{934} ***]
}
\end{minipage}
&
\begin{minipage}[t]{0.23\textwidth}
\centering
\includegraphics[width=.9\textwidth]{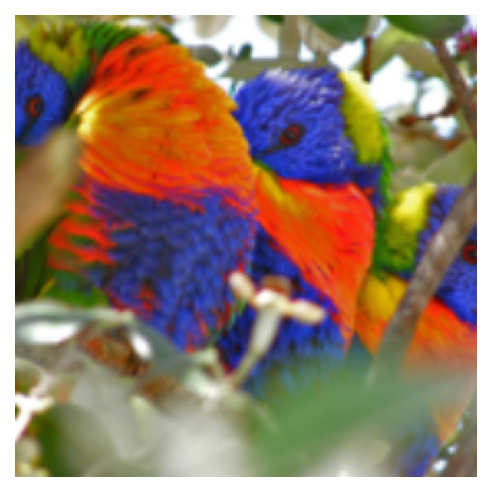} \\
{\scriptsize
\textbf{A:} [** \textcolor{agentagree}{956} \textcolor{classagree}{715} * \textcolor{agentagree}{258} \textcolor{classagree}{834} ***]\\
\textbf{B:} [** \textcolor{agentagree}{956} \textcolor{classagree}{777} * \textcolor{agentagree}{258} \textcolor{classagree}{934} ***]
}
\end{minipage}
&
\begin{minipage}[t]{0.23\textwidth}
\centering
\includegraphics[width=.9\textwidth]{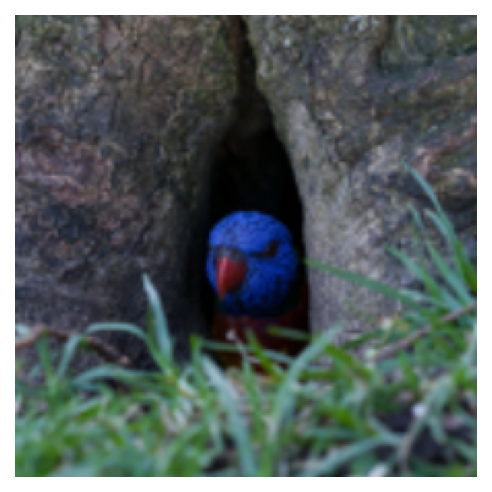} \\
{\scriptsize
\textbf{A:} [** \textcolor{agentagree}{956} \textcolor{classagree}{715} * \textcolor{agentagree}{258} \textcolor{classagree}{834} ***]\\
\textbf{B:} [** \textcolor{agentagree}{956} \textcolor{classagree}{777} * \textcolor{agentagree}{258} \textcolor{classagree}{934} ***]
}
\end{minipage}
&
\begin{minipage}[t]{0.23\textwidth}
\centering
\includegraphics[width=.9\textwidth]{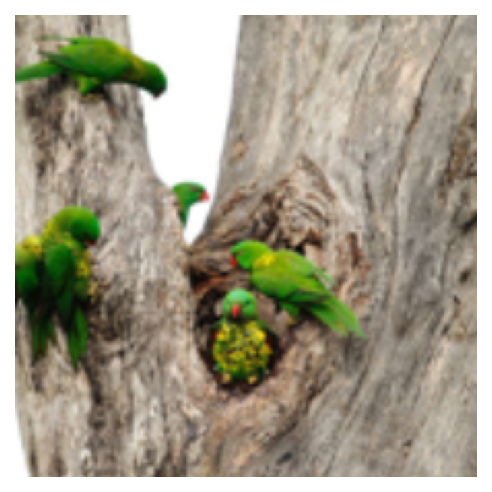} \\
{\scriptsize
\textbf{A:} [** \textcolor{agentagree}{956} \textcolor{classagree}{715} \textcolor{agentagree}{171} \textcolor{agentagree}{258} ****]\\
\textbf{B:} [** \textcolor{agentagree}{956} \textcolor{classagree}{777} \textcolor{agentagree}{171} \textcolor{agentagree}{258} ****]
}
\end{minipage}
\end{tabular}

\caption{{Examples of images and their corresponding emergent signs.}
For each image, we show the discrete messages produced independently by Agent A and Agent B. Red tokens indicate agreement between agents for the same image, while blue tokens indicate tokens consistently reused across images of the same class within an agent.}
\label{fig:imagenet_emlang}
\end{figure}

\paragraph{Qualitative analysis}
To further examine the structure of the learned message space, we visualize message embeddings using t-SNE (t-distributed stochastic neighbor embedding)~\citep{2008_vandermaaten_tsne}. We also analyze discrete messages generated by the two agents on a subset of ImageNet-100 classes. For each class, several images are selected, and the corresponding messages produced independently by Agent A and Agent B are inspected.
\begin{itemize}
\item As shown in the t-SNE visualization of the message space in Fig.~\ref{fig:tsne_messages_cifar10}, SSNG produces a more structured message space with clearer class clusters compared with other communication frameworks. This qualitative observation is consistent with the higher linear-probe accuracy reported in Tables~\ref{tab:exp1_cifar10} and~\ref{tab:exp1_imagenet100}.
\item Figure~\ref{fig:imagenet_emlang} illustrates representative examples from several object categories. Red-highlighted tokens indicate agreement between agents for the same image, while blue-highlighted tokens indicate tokens that are consistently reused across images of the same class within an agent. These examples suggest that SSNG gives rise to emergent signs that are both aligned across agents and structured at the semantic class level.
\item Across the examined classes, we observe consistent cross-agent alignment: for a given image, the two agents generate messages with substantial token overlap despite using heterogeneous perceptual backbones. At the same time, images from the same semantic class exhibit recurring token patterns within each agent, while different classes are associated with largely distinct token patterns.
\end{itemize}
Overall, this qualitative analysis complements the quantitative results by illustrating how SSNG enables agents to develop shared symbolic representations that capture semantic structure in high-dimensional visual domains. Together, these results show that SSNG enables agents to develop a more consistent and aligned communication system than the baselines, particularly on more complex datasets. 


%
%

%
%

%
%

\subsection{Experiment 2: SSNG in Representation Learning} \label{subsec_exp_ssl}
\paragraph{Description}
In the unified single-system configuration, SSNG reduces to a SimSiam-like framework for self-supervised representation learning (Section~\ref{sec: ssng_ssl}). In this setting, the variable $w$ acts as a stochastic bottleneck between representation encoding and prediction. We evaluate whether SSNG can learn transferable visual representations using standard linear evaluation on downstream classification tasks. Experiments are conducted on CIFAR-10 and ImageNet-100. 

\paragraph{Protocol}  
Two augmented views of each input were processed by the backbone–projector $f$, producing projected representations $z_A$ and $z_B$. One branch passed through the predictor $l \circ h$ to obtain $\hat z_A = l(h(z_A))$, which is aligned with $z_B$ under a stop-gradient operation. The roles are swapped to form a symmetric SimSiam-style loss. Models were trained for 200 epochs following standard SSL practice.  

\paragraph{Evaluation}  
We compared SSNG against three baselines—SimSiam, VI-SimSiam, and SimCLR (a non-SimSiam contrastive SSL method)—and included an ablation without the stop-gradient mechanism to test its role in preventing collapse.  
All models shared the same backbone and projector architecture and were trained under identical optimization and seeding conditions.  Following the standard linear evaluation protocol, Top-1 classification accuracy was reported using a linear probe trained on frozen backbone features. Full implementation details and hyperparameters are provided in~\ref{appendix_experiment_settings}.

\paragraph{Results and Discussion}
All experiments were repeated five times with different random seeds, and we report the mean results in Table~\ref{tab:exp2_ssl}, highlight the following findings:

\begin{itemize}
\item \textit{Performance across datasets.} On the simpler CIFAR-10 benchmark, SSNG outperforms all baselines, demonstrating its ability to extract discriminative and transferable features. On the more challenging ImageNet-100 dataset, SSNG performs slightly below SimSiam while still surpassing SimCLR and VI-SimSiam. This minor performance gap suggests that the stochastic bottleneck introduced by $w$ provides useful representational structure but may introduce a trade-off in optimization as data complexity increases. 
\item \textit{Role of stop-gradient.} The stop-gradient mechanism is essential for preventing representational collapse. Removing it causes SSNG to converge to a trivial solution, mirroring the failure mode observed in SimSiam without stop-gradient \citep{2021_chen_simsiam}. This result provides direct evidence that SSNG inherits the core representation-learning of SSL, rather than functioning merely as a communication protocol.
\end{itemize}

Overall, these findings support interpreting SSNG as a principled generalization of SimSiam, preserving its representational learning capabilities while introducing a probabilistic communication bottleneck that supports EmCom.


\begin{table}[t]
\centering
\caption{Top-1 classification accuracy (\%) of different SSL models on CIFAR-10 and ImageNet-100 after linear evaluation. 
All models use the same backbone and projector architecture for fair comparison. Highest values are in bold, and second-highest values are underlined.}
\label{tab:exp2_ssl}
\begin{tabular}{lcc}
\toprule
\textbf{Model} 		& \textbf{CIFAR-10} 				& \textbf{ImageNet-100} \\
\midrule
SimSiam 			& 85.26 {\scriptsize$\pm$0.21} 		& \textbf{79.53} {\scriptsize$\pm$0.33} \\
VI-SimSiam 		& 83.95 {\scriptsize$\pm$0.48} 		& 76.37 {\scriptsize$\pm$0.61} \\
SimCLR 			& \underline{85.49} {\scriptsize$\pm$0.27} 		& 77.14 {\scriptsize$\pm$0.42} \\
SSNG	 		& \textbf{86.10} {\scriptsize$\pm$0.53} & \underline{78.61} {\scriptsize$\pm$0.62}\\
SSNG (no stop-grad) & 10.00 {\scriptsize$\pm$0.00} 	& 1.00 {\scriptsize$\pm$0.00} \\
\bottomrule
\end{tabular}
\end{table}


%
%

\section{Conclusion}

This study introduced the SimSiam Naming Game (SSNG), a framework that connects emergent communication (EmCom) and self-supervised learning (SSL) through a shared objective of latent representation alignment. In SSNG, agents exchange discrete messages generated via a differentiable Gumbel–Softmax mechanism, and each agent updates its internal state by aligning it with the representation inferred from the partner’s message. Inspired by the role of joint attention in human language acquisition, this formulation enables communication systems to emerge without explicit rewards, corrective feedback, or sampling-based update rules. The framework is conceptually related to the Collective Predictive Coding (CPC) hypothesis, as both emphasize the emergence of shared communication structures through interaction rather than explicit supervision. This perspective suggests a possible direction for developing multi-agent systems in which communication protocols arise through representation alignment and interaction.

Experimental results highlight two main findings. First, in the EmCom setting, SSNG produces discrete messages that retain substantially more class-relevant information than those generated by referential games, reconstruction games, and the Metropolis–Hastings Naming Game (MHNG). This advantage persists even when agents employ heterogeneous perceptual backbones, suggesting that representation alignment provides a robust inductive bias for developing shared communication systems. Second, when the two SSNG agents are collapsed into a single network, the resulting model behaves similarly to SimSiam and achieves comparable performance on representation learning tasks, indicating that SSNG preserves the learning dynamics of modern SSL methods.

Despite these promising results, several limitations remain. The current study focuses primarily on image classification datasets, and it remains unclear how SSNG behaves in more complex multimodal or sequential communication environments. In addition, the evaluation of emergent messages relies mainly on linear probing and qualitative analysis; more comprehensive metrics for assessing semantic structure and compositionality would provide deeper insights into the properties of the learned communication systems.

Future work may extend SSNG in several directions, including investigating compositional generalization, scaling the framework to larger populations of interacting agents, and incorporating temporal or sequential communication processes. Integrating richer generative models and multimodal perception modules may also provide a broader testbed for studying how structured symbolic communication emerges in complex environments.

%
%

\bibliographystyle{elsarticle-harv} 
\bibliography{reference.bib}

%
%


\pagebreak
\appendix

%
%

\section{SimSiam Naming Game - Variational Inference Interpretation}
\label{appendix_ssng_objective_function}
\begin{proof}
The objective function of SSNG is derived as follows:
\begin{align}
\mathcal{J}_{\text{SSNG}} 
&:= \mathbb{E}_{q_{\phi}(w, \mathbb{Z} \mid \mathbb{X})} \left[ \log \frac{p_{\theta}(w, \mathbb{Z}, \mathbb{X})}{q_{\phi}(w, \mathbb{Z} \mid \mathbb{X})} \right] \nonumber \\
&= \mathbb{E}_{q_{\phi}(w, \mathbb{Z} \mid \mathbb{X})} \left[ \log p_{\theta}(\mathbb{X} \mid \mathbb{Z}) \right] 
- \mathbb{E}_{q_{\phi}(w, \mathbb{Z} \mid \mathbb{X})} \left[ \log q_{\phi}(\mathbb{Z} \mid \mathbb{X}) \right] + \nonumber \\
&+ \mathbb{E}_{q_{\phi}(w, \mathbb{Z} \mid \mathbb{X})} \left[ \log p_{\theta}(\mathbb{Z} \mid w) \right] 
+ \mathbb{E}_{q_{\phi}(w, \mathbb{Z} \mid \mathbb{X})} \left[ \log p(w) - \log q_{\phi}(w \mid \mathbb{Z}, \mathbb{X}) \right] 
\label{eq: appen-1}
\end{align}

Since $p_{\theta}(\mathbb{X})$ is intractable, we approximate it with empirical data distribution $p_D(\mathbb{X})$. Using Bayes' theorem:
\begin{equation}
p_{\theta}(\mathbb{X} \mid \mathbb{Z}) 
= \frac{p_{\theta}(\mathbb{Z} \mid \mathbb{X}) p_{\theta}(\mathbb{X})}{\mathbb{E}_{p_{\theta}(\mathbb{X})}[p_{\theta}(\mathbb{Z} \mid \mathbb{X})]}
\approx \frac{p_{\theta}(\mathbb{Z} \mid \mathbb{X}) p_D(\mathbb{X})}{\mathbb{E}_{p_D(\mathbb{X})}[p_{\theta}(\mathbb{Z} \mid \mathbb{X})]}
\end{equation}
then
\begin{align}
&\mathbb{E}_{q_{\phi}(w, \mathbb{Z} \mid \mathbb{X})} \left[ \log p_{\theta}(\mathbb{X} \mid \mathbb{Z}) \right] 
\approx \mathbb{E}_{q_{\phi}(w, \mathbb{Z} \mid \mathbb{X})} \left[ \log p_{\theta}(\mathbb{Z} \mid \mathbb{X}) \right] - \mathbb{E}_{q_{\phi}(w, \mathbb{Z} \mid \mathbb{X})} \left[ \log \mathbb{E}_{p_D(\mathbb{X})}[p_{\theta}(\mathbb{Z} \mid \mathbb{X})] \right] + \log p_D(\mathbb{X})
\label{eq: appen-2}
\end{align}

Besides,
\begin{align}
&\mathbb{E}_{q_{\phi}(w, \mathbb{Z} \mid \mathbb{X})} \left[ \log p(w) - \log q_{\phi}(w \mid \mathbb{Z}, \mathbb{X}) \right] 
= - \mathbb{E}_{q_{\phi}(\mathbb{Z} \mid \mathbb{X})} \left[ D_\text{KL} \left( q_{\phi}(w \mid \mathbb{Z}) \parallel p(w) \right) \right]
\label{eq: appen-3}
\end{align}

Substituting Eqs.~(\ref{eq: appen-2}), (\ref{eq: appen-3}) to Eq.~(\ref{eq: appen-1}) and omiting the constants,
\begin{align}
\mathcal{J}_{\text{SSNG}} 
&\approx \mathcal{J}_{\text{align}} + \mathcal{J}_{\text{recon}} + \mathcal{J}_{\text{sign}} + \mathcal{J}_{\text{uniform}} + \log p_D(\mathbb{X}) \nonumber \\
&\propto \mathcal{J}_{\text{align}} + \mathcal{J}_{\text{recon}} + \mathcal{J}_{\text{sign}} + \mathcal{J}_{\text{uniform}}
\end{align}
where
\begin{align}
\mathcal{J}_{\text{align}} &:= \mathbb{E}_{q_{\phi}(\mathbb{Z} \mid \mathbb{X})} \left[ \log p_{\theta}(\mathbb{Z} \mid \mathbb{X}) - \log q_{\phi}(\mathbb{Z} \mid \mathbb{X}) \right] \\
\mathcal{J}_{\text{recon}} &:= \mathbb{E}_{q_{\phi}(w, \mathbb{Z} \mid \mathbb{X})} \left[ \log p_{\theta}(\mathbb{Z} \mid w) \right]  \\
\mathcal{J}_{\text{sign}} &:= - \mathbb{E}_{q_{\phi}(\mathbb{Z} \mid \mathbb{X})} \left[ D_{\text{KL}} \left( q_{\phi}(w \mid \mathbb{Z}) \parallel p(w) \right) \right]  \\
\mathcal{J}_{\text{uniform}} &:= \mathbb{E}_{q_{\phi}(\mathbb{Z} \mid \mathbb{X})} \left[ - \log p_D(\mathbb{Z}) \right] \\
p_D(\mathbb{Z}) &:= \mathbb{E}_{p_D(\mathbb{X})}[p_{\theta}(\mathbb{Z} \mid \mathbb{X})]
\end{align}
\end{proof}

%
%

\section{SimSiam Naming Game - EmCom Training Objectives}
\label{appendix_ssng_elbo_listener}
\begin{proof}
Substituting these definitions in eqs.~\eqref{eq: ssng_z_prior}, \eqref{eq: ssng_w_variational}, \eqref{eq: ssng_z_variational}, \eqref{eq: ssng_z_posterior}, and \eqref{eq: ssng_w_prior} into eqs.~\eqref{eq: loss_ssng_align}, \eqref{eq: loss_ssng_self}, \eqref{eq: loss_ssng_sign}, and \eqref{eq: loss_ssng_uniform}, we derive each component of the ELBO as follows.

\subsubsection*{Alignment Objective}
\begin{align}
\mathcal{J}_{\text{align}} 
&= \mathbb{E}_{q_{\phi}(\mathbb{Z} \mid \mathbb{X})} \left[ \log p_{\theta}(\mathbb{Z} \mid \mathbb{X}) - \log q_{\phi}(\mathbb{Z} \mid \mathbb{X}) \right] \nonumber \\
&= \mathbb{E}_{q_{\phi}(\mathbb{Z} \mid \mathbb{X})} \left[ \log \big( \prod_{i=1}^{M} C_d(\kappa^{(p_z)})\,\exp\!\big(\kappa^{(p_z)}\,\mu_i^{(p_z)\top} z_i^{} \big) \big) - \log \big( \prod_{i=1}^{M} C_d(\kappa^{(q_z)})\,\exp\!\big(\kappa^{(q_z)}\,\mu_i^{(q_z)\top} z_i^{} \big) \big) \right] \nonumber \\
&= \sum_{i=1}^{M} \Big( \kappa^{(p_z)}\,\mu_i^{(p_z)\top} \mathbb{E}_{q_{\phi}(z_i \mid x_i)} \left[ z_i \right] - \kappa^{(q_z)}\,\mu_i^{(q_z)\top} \mathbb{E}_{q_{\phi}(z_i \mid x_i)} \left[ z_i \right] \Big)
+ \sum_{i=1}^{M} \Big( \log C_d(\kappa^{(p_z)}) - \log C_d(\kappa^{(q_z)}) \Big)
\end{align}

\noindent We model the perceptual posterior as a vMF distribution
\begin{align}
    q_{\phi}(z_i \mid x_i) &= \mathrm{vMF}(z_i^{}; \mu_i^{(q_z)} = f_i(x_i), \kappa^{(q_z)}) 
\end{align}
where $\|\mu_i^{(q)}\| = 1$.  
For analytical tractability, we consider the high-concentration regime $\kappa^{(q)} \to \infty$, in which the vMF distribution becomes sharply peaked around its mean direction. This limiting case is used as a mathematical approximation to establish consistency with the operational form of SimSiam (i.e., stop-gradient and cosine similarity maximization), rather than as a literal modeling assumption. In this regime, the expected value satisfies
\begin{equation}
\mathbb{E}_{q_\phi(z_i \mid x_i)}[z_i] \;\approx\; \mu_i^{(q_z)} \;=\; f_i(x_i).
\end{equation}

\noindent Substituting into $\mathcal{J}_{\text{align}}$ gives
\begin{align}
\mathcal{J}_{\text{align}} 
&\approx \sum_{i=1}^{M} \Big( \kappa^{(p_z)}\,\mu_i^{(p_z)\top} \mu_i^{(q_z)} \Big) + \sum_{i=1}^{M} \Big( \log C_d(\kappa^{(p_z)}) - \log C_d(\kappa^{(q_z)}) - \kappa^{(q_z)} \Big)
\label{eq:J_align_1}
\end{align}

\noindent During the optimization, we treat the concentrations $\kappa^{(p_z)}$ and $\kappa^{(q_z)}$  as fixed hyperparameters. Under this assumption, the second sum in \eqref{eq:J_align_1} is constant, Thus, maximizing the $\mathcal{J}_{\text{align}} $ is equivalent to:
\begin{equation}
\mathcal{J}_{\text{align}} \propto \sum_{i=1}^{M} \Big( \mu_i^{(p_z)\top} \mu_i^{(q_z)} \Big) = \sum_{i=1}^{M} g_i(x_j)^{\top}f_i(x_i)
\end{equation}

\subsubsection*{Reconstruction Objective} 
\begin{align}
\mathcal{J}_{\text{recon}} 
&= \mathbb{E}_{q_{\phi}(w, \mathbb{Z} \mid \mathbb{X})} \left[ \log p_{\theta}(\mathbb{Z} \mid w) \right] \nonumber \\
&= \mathbb{E}_{q_{\phi}(\mathbb{Z} \mid \mathbb{X})} \left[ \mathbb{E}_{q_{\phi}(w \mid \mathbb{Z})} \left[ \log p_{\theta}(\mathbb{Z} \mid w) \right] \right]
\end{align}

The inner term \( \mathbb{E}_{q_{\phi}(w \mid \mathbb{Z})} \left[ \log p_{\theta}(\mathbb{Z} \mid w) \right] \) represents the reconstruction from $z \rightarrow w \rightarrow z$. This objective $\mathcal{J}_{\text{recon}}$ measures the alignment between the reconstructed representation $g_\theta(x_i)$ and the original one $f_\phi(x_i)$, and can be approximated as:
\begin{align}
\mathcal{J}_{\text{recon}} \propto \frac{1}{M} \sum_{i=1}^M g_i(x_i)^\top f_i(x_i),
\end{align}

\subsubsection*{Uniform Objective} 
The role of $\mathcal{J}_{\text{uniform}}$ is to ensure that the marginal distribution $p_D(z)$ is uniform over the hypersphere, i.e., $p_D(z) = \mathcal{U}(S^{d-1})$. However, the predictor $h$, defined as a DirectPred \citep{2021_tian_SSL_directpred}, ensures that the latent representations $z$ are uniformly spread over the hypersphere. It achieves this by making the distribution of $z$ approximately isotropic, with each dimension being independent and having equal variance. Consequently, $h$ implicitly maximizes $\mathcal{J}_{\text{uniform}}$ \citep{2023_nakamura_visimsiam}. 

Since the predictor already encourages a uniform distribution of the representations, explicitly including $\mathcal{J}_{\text{uniform}}$ in the total objective is redundant. Therefore, it can be omitted without losing the intended effect on the representation distribution.

\subsubsection*{Sign Objective}
The posterior over messages (eq.~\eqref{eq: ssng_w_variational}) is defined as a mixture of vMF distributions,
\begin{equation}
q_{\phi}(w \mid \mathbb{Z})
= \frac{1}{M} \sum_{i=1}^{M} q_i(w \mid z_i),
\qquad
q_i(w \mid z_i) :=
\mathrm{vMF} \bigl(
w;\, \mu_i^{(w)} = h_i(z_i), \kappa_i^{(w)} \bigr).
\end{equation}

Since the KL divergence $D_{\text{KL}}(q\Vert p)$ is convex in its first argument $q$, Jensen's inequality yields
\begin{align}
D_{\text{KL}}\!
\left( \frac{1}{M} \sum_{i=1}^{M} q_i(w \mid z_i)
\,\Big\Vert\, p(w) \right)
\;\leq\; \frac{1}{M}
\sum_{i=1}^{M} D_{\text{KL}}\!\left(
q_i(w \mid z_i) \,\Vert\, p(w)
\right).
\end{align}

Substituting this into $\mathcal{J}_{\text{sign}}$, we establish a tractable lower bound to maximize: 
\begin{equation}
\mathcal{J}_{\text{sign}} \;\geq\;
- \frac{1}{M} \sum_{i=1}^{M}
\mathbb{E}_{q_{\phi}(\mathbb{Z} \mid \mathbb{X})}
\left[ D_{\text{KL}}\!\left( q_i(w \mid z_i) \,\Vert\, p(w) \right) \right]
\end{equation}

Besides, due to eq.~\eqref{eq: ssng_w_prior}, during communication, the message produced by agent $j$ serves as the prior for agent $i$ :
\begin{align}
p(w) = \mathrm{vMF}\!\left(w; \mu_j = h_j(z_j), \kappa^{(w)} \right), \qquad j \neq i
\end{align}

The inside KL divergence between two vMF distributions can be transformed as follows:
\begin{align}
D_{\text{KL}}\!\left( q_i(w \mid z_i) \,\Vert\, p(w) \right) 
&= \mathbb{E}_{q_i(w \mid z_i)}
\left[ \log q_i(w \mid z_i) - \log p(w) \right] 
=  \kappa^{(w)} \mu_i^{(w)\top} \mathbb{E}_{q_i(w \mid z_i)} \left[ w \right] 
- \kappa^{(w)} \mu_j^\top \mathbb{E}_{q_i(w \mid z_i)} \left[  w \right] 
\label{eq: app_kl_expanded}
\end{align}

We assume a high-concentration regime that $\kappa^{(w)} \to \infty$, under which the vMF distribution converges to a point mass at its mean direction, then
\begin{equation}
\mathbb{E}_{q_i(w \mid z_i)} \left[  w \right] \;\approx\; \mu_i^{(w)} \;=\; h_i(z_i)
\end{equation}
Under these assumptions, \eqref{eq: app_kl_expanded} simplifies to
\begin{equation}
D_{\text{KL}}\!\left( q_i(w \mid z_i) \,\Vert\, p(w) \right)  \;\approx\;
- \kappa^{(w)} \mu_j^\top \mu_i^{(w)}
\end{equation}

Substituting the above approximation into the lower bound of
$\mathcal{J}_{\text{sign}}$ and treating the concentration parameters as fixed
hyperparameters, we obtain
\begin{align}
\mathcal{J}_{\text{sign}}
\;\gtrsim\;
\frac{\kappa^{(w)}}{M}
\sum_{i=1}^{M}
\mathbb{E}_{q_{\phi}(\mathbb{Z} \mid \mathbb{X})}
\!\left[
\mu_j^\top \mu_i^{(w)}
\right]
\;+\;\text{const},
\end{align}
where the constant term does not depend on the learnable parameters and is therefore omitted. Using the definitions $\mu_i^{(w)} = h_i(z_i)$ and
$\mu_j = h_j(z_j)$, this yields
\begin{align}
\mathcal{J}_{\text{sign}}
\;\propto\;
\sum_{i=1}^{M}
\mathbb{E}_{q_{\phi}(\mathbb{Z} \mid \mathbb{X})}
\!\left[
h_j(z_j)^\top h_i(z_i)
\right].
\end{align}

In practice, the expectation over $q_{\phi}(\mathbb{Z} \mid \mathbb{X})$ is
estimated using a single Monte Carlo sample per mini-batch, resulting in the
empirical objective
\begin{align}
\mathcal{J}_{\text{sign}}
\;\propto\;
\sum_{i=1}^{M}
h_j(z_j)^\top h_i(z_i).
\end{align}

\subsubsection*{Total Objective}
Combining these components, the objective function of SSNG is given by:
\begin{align}
\mathcal{J}_{\text{SSNG}}
\propto \alpha \sum_{i=1}^{M} g_i(x_j)^\top f_i(x_i)
\;+\; \beta \sum_{i=1}^{M} g_i(x_i)^\top f_i(x_i)
\;+\; \gamma \sum_{i=1}^{M} h_j(z_j)^{\top} h_i(z_i) 
\end{align}
where $\alpha$, $\beta$, and $\gamma$ are scalar hyperparameters that control the relative weights of the three terms, and $j \neq i$ denotes the agent from which agent $i$ receives a message.

Substituting eqs.~\eqref{eq: agent_perc_enc}, \eqref{eq: agent_lang_enc}, and \eqref{eq: agent_lang_dec} yields the equivalent form
\begin{align}
\mathcal{J}_{\text{SSNG}}
= \alpha \sum_{i=1}^{M} (z_i^{(w_j)})^\top (z_i^{(x_i)})
\;+\; \beta \sum_{i=1}^{M} (z_i^{(w_i)})^\top (z_i^{(x_i)})
\;+\; \gamma \sum_{i=1}^{M} (w_j)^{\top} (w_i) 
\label{eq: app_ssng_total}
\end{align}

This global objective is carried out for both agents. The objective for agent $i$ is then defined as
\begin{align}
\mathcal{J}_i
:= \alpha \,\, ({z}_i^{(w_j)})^\top (z_i^{(x_i)})
\;+\; \beta \,\, ({z}_i^{(w_i)})^\top (z_i^{(x_i)}) 
\;+\; \gamma \,\, (w_j)^\top (w_i)
\end{align}
\end{proof}

%
%

\section{SimSiam Naming Game - Representation Learning}
\label{appendix_ssng_elbo_representation}
\begin{proof}
The objective function of the SSNG model in representation learning is derived from:
\begin{align}
\mathcal{L}_{\text{SSL}} := \mathbb{E}_{q_{\phi}(z, w | \mathbb{X})} \left[ \log \frac{p_{\theta}(w, z, \mathbb{X})}{q_{\phi}(z, w | \mathbb{X})} \right]
\end{align}
Similar to the previous proof, the objective function is:
\begin{align}
\mathcal{J}_{\text{SSL}} := \mathcal{J}_{\text{align}} + \mathcal{J}_{\text{recon}} + \mathcal{J}_{\text{sign}} + \mathcal{J}_{\text{uniform}} 
\end{align}
where
\begin{align}
\mathcal{J}_{\text{align}} &:= \mathbb{E}_{q_{\phi}(z | \mathbb{X})} \left[ \log p_{\theta}(z | \mathbb{X}) - \log q_{\phi}(z | \mathbb{X}) \right] \\
\mathcal{J}_{\text{recon}} &:= \mathbb{E}_{q_{\phi}(z, w | \mathbb{X})} \left[ \log p_{\theta}(z | w) \right]  \\
\mathcal{J}_{\text{sign}} &:= - \mathbb{E}_{q_{\phi}(z | \mathbb{X})} \left[ D_{\text{KL}} \left( q_{\phi}(w | z, \mathbb{X}) \| p(w) \right) \right]  \\
\mathcal{J}_{\text{uniform}} &:= \mathbb{E}_{q_{\phi}(z | \mathbb{X})} \left[ - \log p_D(z) \right] \\
p_D(z) &:= \mathbb{E}_{p_D(\mathbb{X})}[p_{\theta}(z | \mathbb{X})]
\end{align}

Using the definitions in eqs.~\eqref{eq: ssl_w_prior}, \eqref{eq: ssl_z_prior}, \eqref{eq: ssl_w_variational}, \eqref{eq: ssl_z_variational}, \eqref{eq: ssl_z_posterior}:

\subsubsection*{Alignment Objective}
\begin{align}
\mathcal{J}_{\text{align}} = \mathbb{E}_{q_{\phi}(z | \mathbb{X})} \left[ \log p_{\theta}(z | \mathbb{X}) \right] - \mathbb{E}_{q_{\phi}(z | \mathbb{X})} \left[ \log q_{\phi}(z | \mathbb{X}) \right] \nonumber
\end{align}

\noindent We approximate $\mathbb{E}_{q_\phi(z \mid \mathbb{X})}\!\left[ \log q_\phi(z \mid \mathbb{X}) \right]$ using Monte Carlo sampling by drawing $K$ i.i.d. samples
\[
z_k \sim q_\phi(z \mid \mathbb{X}), \qquad k = 1,\dots,K,
\]
by first sampling a component index $I_k \sim \mathrm{Unif}\{1,\dots,M\}$ and then sampling $z_k \sim \mathrm{vMF}(z;\mu_{I_k},\kappa)$.
The objective is approximated as
\begin{align}
\mathbb{E}_{q_\phi(z \mid \mathbb{X})}\!\left[ \log q_\phi(z \mid \mathbb{X}) \right]
\approx -\frac{1}{K}\sum_{k=1}^{K} \log q_\phi(z_k \mid \mathbb{X}) 
= -\frac{1}{K}\sum_{k=1}^{K}  \log \left( \frac{1}{M} \sum_{i=1}^{M} \text{vMF}(z_k; \mu_i^{(q_z)}, \kappa^{(q_z)}) \right) \nonumber
\end{align}

\begin{align}
\mathcal{J}_{\text{align}} 
&\approx \mathbb{E}_{q_{\phi}(z | \mathbb{X})} \left[ \log \left( \prod_{i=1}^{M} \text{vMF}(z; \mu_i^{(p_z)}, \kappa^{(p_z)}) \right) \right] 
-\frac{1}{K}\sum_{k=1}^{K}  \log \left( \frac{1}{M} \sum_{i=1}^{M} \text{vMF}(z_k; \mu_i^{(q_z)}, \kappa^{(q_z)}) \right)  \nonumber \\
&= \sum_{i=1}^{M} \Big( \kappa^{(p_z)}\,\mu_i^{(p_z)\top} \mathbb{E}_{q_{\phi}(z \mid x_i)} \left[ z \right] \Big)
- \frac{1}{K}\sum_{k=1}^{K}  \log \left( \sum_{i=1}^{M} \exp \kappa^{(q_z)}\,\mu_i^{(q_z)\top} z_k \right) \nonumber \\
&+ \sum_{i=1}^{M} \Big( \log C_d(\kappa^{(p_z)}) \Big) 
- \frac{1}{K}\sum_{k=1}^{K}  \log \left( \frac{C_d(\kappa^{(q_z)})}{M} \right)  
\end{align}

We consider the high-concentration regime $\kappa^{(q_z)} \to \infty$, in which the posterior becomes sharply peaked and stochasticity vanishes. This limiting case is introduced as a mathematical approximation to establish consistency with the deterministic operations used in SimSiam, rather than as a literal modeling assumption.
\begin{equation}
\lim_{\kappa^{(q_z)} \to \infty} z_k = \mu_i^{(q_z)}, \qquad \text{and} \qquad \mathbb{E}_{q_{\phi}(z \mid x_i)} [z] \approx \mu_i^{(q_z)}
\end{equation}

Hence,
\begin{align}
\mathcal{J}_{\text{align}} 
&= \sum_{i=1}^{M} \Big( \kappa^{(p_z)}\,\mu_i^{(p_z)\top} \mu_i^{(q_z)} \Big)
- \frac{1}{K}\sum_{k=1}^{K}  \log \left( \sum_{i=1}^{M} \exp \kappa^{(q_z)}\,\mu_i^{(q_z)\top} \mu_i^{(q_z)} \right) \nonumber \\
&+ \sum_{i=1}^{M} \Big( \log C_d(\kappa^{(p_z)}) \Big) 
- \frac{1}{K}\sum_{k=1}^{K}  \log \left( \frac{C_d(\kappa^{(q_z)})}{M} \right)  
\end{align}

During the optimization, we treat the concentrations as constant, we obtain 
\begin{align}
\mathcal{J}_{\text{align}}
\propto \sum_{i=1}^{M} \mu_i^{(p_z)\top} \mu_i^{(q_z)}  = \sum_{i=1}^{M} g(x_j)^{\top} f(x_i), \qquad \text{where } j \neq i
\end{align}

\subsubsection*{Reconstruction Objective}
The reconstruction loss $\mathcal{J}_{\text{recon}}$ measures the alignment between the reconstructed representation $g(x_i)$ and the original one $f(x_i)$. 
\begin{align} 
\mathcal{J}_{\text{recon}} \approx \sum_{i=1}^{M}  g(x_i)^\top f(x_i) 
\end{align}
This alignment is already captured by the $\mathcal{J}_{\text{align}}$. Hence, $\mathcal{J}_{\text{recon}}$ is omitted from the total loss.

\subsubsection*{Uniform Objective} 
Similar to original SSNG, $\mathcal{J}_{\text{uniform}}$ can be omitted since the predictor already encourages a uniform distribution of the representations.

\subsubsection*{Sign Objective}
Because the stochastic bottleneck $w$ is generated from the latent representation $z_i$, the sign regularization term can be written as
\begin{align}
\mathcal{J}_{\text{sign}} 
&\propto - \sum_{i=1}^{M}  D_{\text{KL}} \left( q_{\phi}(w | z_i) \| p(w) \right)
\end{align}

\subsubsection*{Total Objective}
Let $x_A$ and $x_B$ denote two augmented views of the same input $x$, then
\begin{align}
z_A &= f(x_A), \quad \hat z_A = g(x_A) = l( h( z_A )) \nonumber \\
z_B &= f(x_B), \quad \hat z_B = g(x_B) = l( h( z_B )) \nonumber 
\end{align}

The total objective for SSL is defined as
\begin{align}
\mathcal{J}_{\mathrm{SSL}} &:= \mathcal{J}_{\text{align}} - \beta_{\mathrm{KL}} \, \mathcal{J}_{\mathrm{KL}} 
\end{align}
where
\begin{align}
\mathcal{J}_{\mathrm{align}}
&:= (\hat z_A)^{\top}(z_B) + (\hat z_B)^{\top}(z_A), \\
\mathcal{J}_{\mathrm{KL}} 
&:= D_{\mathrm{KL}} \Big( q_{\phi}(w \mid z_A) \;\big\|\; p(w) \Big) + D_{\mathrm{KL}} \Big( q_{\phi}(w \mid z_B) \;\big\|\; p(w) \Big)
\end{align}
\end{proof}

%
%

\pagebreak
\section{Experiment Settings}
\label{appendix_experiment_settings}
\begin{table}[!htbp]
\centering
\caption{Experimental settings for the emergent communication (SSNG) experiments.}
\label{tab:exp_emcom}
\begin{tabular}{lcc}
\hline
\textbf{Setting} & \textbf{CIFAR-10} & \textbf{ImageNet100} \\
\hline
Dataset & CIFAR-10 & ImageNet100 \\
Image size & 224 $\times$ 224 & 224 $\times$ 224 \\
Backbone (Agent A) & ResNet-18 & ResNet-18 \\
Backbone (Agent B) & ResNet-34 & ResNet-34 \\
Feature dimension & 512 & 512 \\
Latent dimension & 2048 & 2048 \\
Message length & 10 & 10 \\
Dictionary size & 100 & 1000 \\
Batch size & 256 & 128 \\
Learning rate & 0.05 & 0.02 \\
Optimizer & SGD (momentum 0.9) & SGD (momentum 0.9) \\
Weight decay & $1\times10^{-4}$ & $1\times10^{-4}$ \\
Weights $(\alpha,\beta,\gamma)$ & (1.0, 1.0, 1.0) & (0.3, 0.3, 0.4) \\
Message discretization & Gumbel-Softmax + STE & Gumbel-Softmax + STE \\
Temperature annealing & Exponential decay & Exponential decay \\
Warm-up epochs (soft tokens) & 20 & 20 \\
Training epochs & 100 & 100 \\
\hline
\end{tabular}
\end{table}

\begin{table}[!htbp]
\centering
\caption{Experimental settings for the self-supervised representation learning (SSL) experiments.}
\label{tab:exp_ssl}
\begin{tabular}{lcc}
\hline
\textbf{Setting} & \textbf{CIFAR-10} & \textbf{ImageNet100} \\
\hline
Dataset & CIFAR-10 & ImageNet100 \\
Image size & 224 $\times$ 224 & 224 $\times$ 224 \\
Backbone & ResNet-18 & ResNet-18 \\
Feature dimension & 512 & 512 \\
Latent dimension & 2048 & 2048 \\
Batch size & 256 & 128 \\
Learning rate & 0.05 & 0.02 \\
Optimizer & SGD (momentum 0.9) & SGD (momentum 0.9) \\
Weight decay & $1\times10^{-4}$ & $1\times10^{-4}$ \\
Weights $(\beta_{reg})$ & $1\times10^{-3}$ & $1\times10^{-3}$ \\
Training epochs & 100 & 100 \\
\hline
\end{tabular}
\end{table}

%
%

\begin{sidewaystable}[t]
\section{Comparison of Emergent Communication Frameworks}
\label{appendix_emcom_comparison}
\centering
\begin{tabular}{|p{3cm}|p{4cm}|p{4cm}|p{4cm}|p{4cm}|}
\hline
\textbf{Aspect} & \textbf{SimSiam Naming Game} & \textbf{Metropolis--Hastings (MH) Naming Game} & \textbf{Referential Game} & \textbf{Reconstruction Game} \\
\hline

Learning Paradigm 
& Self-supervised representation learning.
& Unsupervised representation learning.
& Discriminative signal transmission. 
& Reconstructive signal transmission. \\
\hline

Objective 
& Internal representation alignment across agents.
& Probabilistic symbol updates via MCMC.
& Referring to target objects in a shared context. 
& Encoding information for reconstructing the observed object. \\
\hline

Communication Signal
& Agents exchange messages generated from internal representations. 
& Agents propose and iteratively update messages using an MH-based acceptance criterion. 
& Speaker sends a message describing a target object. 
& Speaker sends a message used by the listener to reconstruct the object. \\
\hline

Training Objective
& Maximize representation similarity between agents through exchanged messages.
& Maximize the ELBO conditioned on symbols negotiated via MH sampling.
& Maximize correct identification of the target object among distractors.
& Minimize reconstruction error between original and predicted observations. \\
\hline

Agent Interaction 
& Parallel symmetric interaction with simultaneous message exchange. 
& Sequential turn-taking interaction between agents. 
& One-way: speaker produces a message, listener interprets it. 
& One-way: speaker sends message, listener reconstructs the observation. \\
\hline

Agent Roles 
& Symmetric agents exchange messages and align representations. 
& Agents alternate roles during MH updates. 
& Fixed speaker--listener roles. 
& Fixed encoder--decoder roles. \\
\hline

Observations 
& Different augmented views of the same object. 
& Different augmented views of the same object. 
& Single view of target object alongside distractor objects. 
& Single view of the object to be reconstructed. \\
\hline

Internal Representations
& Shaped to be invariant across different views and agents via non-contrastive alignment.
& Shaped to be {probabilistic explanations} of both observations and symbols.
& Shaped to encode {discriminative features} for target identification. 
& Shaped to encode {fine-grained details} for observation reconstruction. \\
\hline

\end{tabular}
\caption{Comparison of four emergent communication frameworks: SimSiam Naming Game - this work, Metropolis--Hastings Naming Game~\citep{2023_taniguchi_inter_gmm_vae}, Referential Game~\citep{2017_lazaridou_referential}, and Reconstruction Game~\citep{2020_chaabouni_recon_compositionality_generalization}.}
\label{tab: EmCom_game_comparison}
\end{sidewaystable}

%
%

\end{document}